\documentclass[twoside,leqno,twocolumn]{article}
\usepackage{ltexpprt}

\usepackage{booktabs} 
\usepackage{subfigure}
\usepackage{graphicx}
\usepackage{epstopdf}
\usepackage{enumitem}
\usepackage{amssymb}
\usepackage{amsmath}
\usepackage[hyphens]{url}

\usepackage{algorithm}
\usepackage{algorithmic}

\def \v {\mathbf{v}}
\def \w  {\mathbf{w}}

\def \x {\mathbf{x}}

\def \I {\mathbb{I}}

\def \R {\mathbb{R}}

\def \V {\mathbf{V}}
\def \Y {\mathcal{Y}}

\def \bq  {\begin{eqnarray}}
\def \eq  {\end{eqnarray}}
\def \bqs {\begin{eqnarray*}}
	\def \eqs {\end{eqnarray*}}

\begin{document}



\title{\Large A Boosting Framework of Factorization Machine}

\author{Longfei Li\thanks{Ant Financial} \\
	\and
	Peilin Zhao\thanks{Ant Financial} \\
	\and
	Chaochao Chen \thanks{Ant Financial}\\
	\and
	Jun Zhou \thanks{Ant Financial}
	\and 
	Xiaolong Li \thanks{Ant Financial}
	}
\date{}

\maketitle
\fancyfoot[R]{\footnotesize{\textbf{Copyright \textcopyright\ 2018 by SIAM\\
			Unauthorized reproduction of this article is prohibited}}}

\begin{abstract}
	Recently, Factorization Machines (FM) has become more and more popular for recommendation systems, due to its effectiveness in finding informative interactions between features. Usually, the weights for the interactions is learnt as a low rank weight matrix, which is formulated as an inner product of two low rank matrices. This low rank can help improve the generalization ability of Factorization Machines. However, to choose the rank properly, it usually needs to run the algorithm for many times using different ranks, which clearly is inefficient for some large-scale datasets. To alleviate this issue, we propose an Adaptive Boosting framework of Factorization Machines (AdaFM), which can adaptively search for proper ranks for different datasets without re-training. Instead of using a fixed rank for FM, the proposed algorithm will adaptively gradually increases its rank according to its performance until the performance does not grow, using boosting strategy. To verify the performance of our proposed framework, we conduct an extensive set of experiments on many real-world datasets.  Encouraging empirical results shows that the proposed algorithms are generally more effective than state-of-the-art other Factorization Machines.
\end{abstract}

\section{Introduction}
Originally introduced in~\cite{DBLP:conf/icdm/Rendle10}, the Factorization Machines (FM) is proposed as a new model class that combines the advantages of linear models, such as Support Vector Machines (SVM)~\cite{DBLP:books/daglib/0026018}, with factorization models. Like linear model, FM is a general model which will learn a weight vector for any real valued feature vector. However, FM also learn a pairwise feature interaction matrix for all interactions between variables, thus it can estimate interactions for highly sparse data(like recommender systems) where linear models fail.The interaction matrix is learnt using factorized parameters with much smaller latent factor compared with the original dimension of the instances. This introduced several benefits. Firstly, this acts as a kind of regularization, since the rank of the interaction matrix is no more than the latent factors and the number of parameters is much lower than that of the full matrix. Secondly, this makes the computation of the prediction score of FM  can be calculated in linear time and thus FMs can be optimized directly. Because of these advantages, FM can be used for any supervised learning tasks, including classification, regression, and recommendation systems. On the other hand, FM can mimic most factorization models~\cite{DBLP:conf/sigir/RendleGFS11,DBLP:journals/tist/Rendle12}, including standard matrix factorization~\cite{DBLP:conf/nips/SrebroRJ04}, SVD++~\cite{DBLP:conf/kdd/Koren08}, timeSVD++~\cite{DBLP:journals/cacm/Koren10}, and PITF (Pairwise Interaction Tensor Factorization)~\cite{DBLP:conf/wsdm/RendleS10}, just by feature engineering. This property makes FM suitable to many application domains, where factorization models are appropriate. Practically, FM can achieve as good accuracy performance as the best specialized models on the Netflix and KDDcup 2012 challenges~\cite{DBLP:journals/pvldb/Rendle13}.

Although the original Factorization machine is successfully applied to optimize the accuracy of the model~\cite{DBLP:conf/icdm/Rendle10}.However, it is not guaranteed to optimize ranking performance for recommendation system~\cite{DBLP:conf/nips/CortesM03,DBLP:conf/recsys/CremonesiKT10}. Recently the Pair-wised Ranking based Factorization Machines (PRFM) algorithm~\cite{DBLP:conf/cikm/QiangLY13} is proposed to directly optimize the Area Under the ROC Curve (AUC) performance. However, AUC measure is not suitable for top-N recommendation tasks~\cite{DBLP:conf/icml/McFeeL10}, where the higher accuracy at the top of the list is more important than that at the low-position ( such as Normalized Discounted Cumulative Gain (NDCG) and Mean Reciprocal Rank (MRR)~\cite{DBLP:books/daglib/0027504}). So, LambdaFM~\cite{DBLP:conf/cikm/YuanGJCYZ16} is proposed to directly optimize the rank biased metrics, using the core ideas of LambdaRank~\cite{DBLP:conf/nips/BurgesRL06} where top pairs are assigned with higher importance. Empirical results show that LambdaFM generally outperforms PRFM in terms of different ranking metrics. Although FM and their variants are successfully applied to many problems, it usually needs to run the algorithm for many times to choose the rank properly. This clearly is inefficient for some large-scale datasets.

Motivated by the above observations, we would like to design an algorithm that can adaptively search for a proper latent number for different datasets without re-training. To achieve this goal, we adopt boosting technique, which was proposed to improve the performance (such as, AUC, NDCG, MRR) of models by combining multiple weak models~\cite{freund1995desicion,xu2007adarank}, to propose an Adaptive boosting framework of Factorization Machine (AdaFM). Specifically, AdaFM works in rounds to build multiple component FMs bases on dynamically weighted training datasets, which are linearly combined to construct a strong FM. In this way, AdaFM will adaptively gradually increases its latent number according to its performance until the performance becomes saturated. As for component FM, we can either choose the original FM, PRMF or LambdaFM, according the performance that we would like to optimize. To verify the performance of our proposed framework, we conduct an extensive set of experiments on many large-scale real-world datasets.  Encouraging empirical results shows that the proposed algorithms are  more effective than state-of-the-art other Factorization Machines.

The rest of the paper is organized as follows. Section 2 presents the proposed framework and algorithms. Section 3 discusses our experimental results and Section 4 concludes our work.

\section{Adaptive Boosting Factorization Machine}
In this section, we will firstly introduce the problem setting and Factorization Machine. Then, we will present our Adaptive Boosting Factorization Machine framework, following which we will give several specific algorithms.
\subsection{Problem Settings}
Our goal is to learn a function $f: \R^d\rightarrow \Y$, based on a dataset $\Big\{(\x_i, y_i)|i\in[n]:=\{1,\ldots,n\}\Big\}$, where $\x_i\in\R^d$ is the feature vector of the $i$-th instance, $y_i\in\Y$ is the label of $\x_i$. There are many different choices of $\Y$, which corresponds to different problems. For example, when $\Y=\{-1,+1\}$, we can treat this problem as a classification problem.

\subsubsection{Factorization Machine}
To learn a reasonable $f$, Factorization Machines (FM) can be adopted. Specifically, \emph{second order} FM model predict the output for an instance $\x$ using the following simple equation as:
\bq
f_\Theta(\x) = \w^\top\x+ \sum^d_{l=1}\sum^d_{m=l+1}(\V\V^\top)_{lm} x_l x_m
\eq
where $x_l$ is the $l$-th element of $\x$, and the model parameters $\Theta$ to be learnt consists
\bqs
\w\in\R^d,\quad \V=[\v_1,\ldots,\v_k]\in \R^{d\times k}
\eqs
where $k\ll d$ is  a usually prefixed parameter which defines the rank of the factorization.

Intuitively, the vector $\w$, the linear part of the  model, contains the weights of individual features for predicting $y$; while the  positive semidefinite matrix $\V\V^\top$, the factorization part, captures all the pairwise interactions between all the variables. Using the factorized parametrization $\V\V^\top$ instead of a full matrix is based on the assumption that the effect of pairwise interactions has a low rank. This explicit low rank assumption helps reduce the overfitting problem, and allows FM to estimate reliable parameters even in highly sparse data. In addition, this reduces the number of parameters to be learnt from $d^2$ to $kd$, and allows to compute prediction efficiently by using
\bq
\begin{split}
f_\Theta(\x) = \w^\top\x + \frac{1}{2}(\|\V^\top\x\|^2_2- \\
\sum^k_{s=1}\|\v_s\circ\x\|^2_2)
\end{split}
\eq
where $\circ$ is the element-wise product. So, FM can be computed efficiently with the computation cost $O(kd)$ instead of $O(d^2)$ when implemented naively.

Given the above parametric FM function, now we take $\Y=\{-1,+1\}$ as a concrete example, which can be treated as a classification problem. In order to learn the optimal parameters for FM, we need introduce some loss function $\ell(f_\Theta(\x_i),y_i)$ to measure the performance of $f_\Theta$ on $(\x_i,y_i)$. One popular loss function is the well-known logistic regression loss,
\bqs
\ell(f_\Theta(\x_i),y_i)=\ln\big(1+\exp(-y_if_\Theta(\x_i))\big),
\eqs
which measures how much is violation of the desired constraint $y_if_\Theta(\x_i)\ge 0$ by the function $f$. Under these settings, FM is formulated as
\bq
\min_\Theta \frac{1}{n}\sum^n_{i=1}\ell(f_\Theta(\x_i),y_i)+\frac{\gamma}{2}\|\Theta\|^2
\eq
where $\|\Theta\|^2=\|\w\|^2_2 + \|\V\|^2_F$. The parameter $\gamma>0$ is a trade-off parameter for the regularization and empirical loss.



\subsubsection{Pairwise Ranking Factorization Machine}
Although traditional FM can be applied to many different problems with interactions hard to be estimated, it is usually designed to approximately minimize the classification error, or regression loss, which is apparently not appropriate for ranking tasks where the prediction score does not matters while the ranks matter.


To solve this task, Pairwise Ranking Factorization Machines (PRFM) is proposed. In PRFM, the dataset is firstly transformed to a new one which is
\bqs
\{(\x_i,\x_j, y_{ij})|i,j\in[n]\},
\eqs
where $y_{ij}=1$ if $y_i\succ y_j$ and $y_{ij}=-1$ otherwise. Then the objective function of PRFM is defined as
\bq
\begin{split}
	\min_\Theta\frac{1}{n^2}\sum^n_{i,j=1}\ell(f_\Theta(\x_i)-f_\Theta(\x_j),y_{ij})+\\
	\frac{\gamma}{2}\|\Theta\|^2
\end{split}
\eq

where $\ell$ is the logistic regression loss, and $\gamma>0$ is a regularization parameter. Intuitively, PRFM model would assign higher sores for  positive instances compared with negative instances, which is equivalent to approximately maximize a concave lower bound of AUC performance measure. In practice, PRFM dose work much better than FM in the setting of recommendation task measured by AUC.

\subsubsection{Lambda Factorization Machine}
\label{sec:lfm}
Although PRFM can achieve significant higher AUC performance compared with traditional FM. However, in PRMF, an incorrect pairwise ordering at the bottom of list impacts the score just as much as that at the top of the list, this makes it not suitable to top-N recommendation tasks , where the higher accuracy at the top of the list is more important to the recommendation quality than that at the low-position. This can be further
explained using rank biased metrics, such as NDCG and MRR [12] , for which higher weights are assigned to the top accurate instances.

To address this issue, LambdaFM is proposed to directly optimize the rank biased metrics, using the core ideas of LambdaRank where different pairs are assigned with different importance according to their positions in the list. Specifically, three strategies are proposed in LambdaFM. The first one is Static Sampler, in which the item $\x_j$ is assigned to a sampling probability
\bq~\label{ss}
\exp[-(r(\x_j)+1)/(|I|\times\rho)],\quad \rho\in(0,1],
\eq
where $r(\x_j)$ represents the rank of item $\x_j$ among all items $I$ according to its overall popularity, $\rho>0$ is a parameter. The second one is Dynamic Sampler. Dynamic sampler will first draw samples $\x_{j_1},\ldots,\x_{j_m}$ uniformly from unobserved item set $I\backslash I_u,$
where $I_u$ is the item set clicked by $u$, then sample one item according to the distribution
\bq
\begin{split}
	p_j\propto \exp\left[-(r(\x_{j})+1)/(m\rho)\right], \\
	\quad\textrm{where } r(\x_{j_m})\propto 1/\hat{y}(\x_{j_m}),
\end{split}
\eq
where $\hat{y}(\x)=f_\Theta(\x)$. Different from the first two samplers which would like to push non-positive items with higher ranks down from top positions, the third one is to pull positive items with lower ranks up from the bottom positions. Specifically, for a pair of positive and non-positive items $(\x_i,\x_j)$, a rank-aware weight will be assigned to it, where the weight is
\bq
\begin{split}
	\Gamma(r(\x_i))=(\sum^{r(\x_i)}_{r=0}1/(r+1))/\Gamma(I),\\
	\quad \Gamma(I)=\sum^{|I|}_{r=0}1/(r+1).
\end{split}
\eq
However, it is impractical to compute $r(\x_i)$ for large scale datasets. To remedy this issue, an approximate method is to repeatedly draw an item from $I$ until we obtain $\x_j$, s.t., $\hat{y}(\x_i)-\hat{y}(\x_j)\le\epsilon$ and $y_i\succeq y_j$, where $\epsilon$ is a positive margin value. Let $T$ denote the size of sampling trials before obtaining such an item, then $\Gamma(r(\x_i))\approx\lceil\frac{|I|-1}{T}\rceil$. Empirical results show that the three variant of LambdaFM generally outperforms PRFM in terms of different ranking metrics, such as NDCG.

\subsection{Algorithm}
The proposed Adaptive Boosting Factorization Machine (AdaFM) framework aims to provide a general framework to optimize the loss
function defined based on various ranking metrics.

To introduce the proposed algorithm, we briefly describe the problem details with some notations. Specifically, let $U$ be the whole set of useres and $I$ the whole set of items, then our goal is to utilize the interactions between $U$ and $I$ to recommend a target user $u$ a list of items that he may prefer. In training, a set of user $U=\{u_a|a=1,\ldots,n\}$ is given. Each user $u_a$ is associated with a list of retrieved items $I_a=\{i_{ab}|b=1,\ldots,n_a\}$  and a list of labels $Y_a=\{y_{ab}|b=1,\ldots,n_a\}$, where $y_{ab}$ denotes the rank of item $i_{ab}$ for user $u_a$. A feature vector $\x_{ab}$ is created from each user-item pair $(u_a,i_{ab})$. The interaction $y_{ab}$ belongs to the set of $\Y=\{r_1,\ldots,r_q\}$. Thus the training set can be represented as $S=\{(u_a,I_a,Y_a)\}$.  For a user $u_a$ and item $i_{ab}$, we denote his historical items by $I_{ab}=\{i_{ac}\in I_a|y_{ac}=y_{ab}\}$ and define $I^-_{ab}=\{i_{ac}\in I_a|y_{ac}\prec y_{ab}\}$.

\subsubsection{AdaFM}
Our objective is to learn a Factorization Machine $f$, such that for each user $u_a$ the function $f$ can assign its item list $I_a$ with prediction scores that generate a rank list as close as possible with $Y_a$. To achieve this goal, we introduce function $\pi(u_a,I_a,f)$ to denote the rank list of items $I_a$ for $u_a$, resulted by the learnt model $f$. Specifically, for $I_a=\{i_{a1},\ldots,i_{an_a}\}$, $\pi(u_a,I_a,f)$ is defined as a bijection from $\{1,\ldots,n_a\}$ to itself, where the $b$-th element of  $\pi(u_a,I_a,f)$ denotes the rank of item $i_{ab}\in I_a$.

\setlength{\intextsep}{0pt}
\begin{algorithm}[htpb]
	\caption{The Component Algorithm} \label{alg:CA}
	\begin{algorithmic}
		\STATE {\bf Input}: The observed dataset $S=\{(u_a,I_a,Y_a)\}$, the weight distribution $\{p^t_a\}$, the learning rate $\eta$, and the regularization parameter $\gamma$
		\STATE {\bf Initialize:} $w_l=0$, and $v_{l,m}$ using $\mathcal{N}(0,0.1)$
		\FOR{$e=2,\ldots,MaxIter$}
		\STATE Uniformly draw $u_a$ from $U$
		\STATE Uniformly draw $i_{ab}$ from $I_{ab}$
		\STATE Several methods to draw $i_{ac}$ from $I_{ab}^-$:
		\begin{itemize}
			\item Uniformly draw $i_{ac}$   [PRFM]
			\item Randomly draw $i_{ac}$ by Eq~\eqref{ss} [LFM-S]
			\item Randomly draw $i_{ac}$ by Eq~\eqref{cs} [LFM-D]
			\item Randomly draw $i_{ac}$ by Eq~\eqref{rs} [LFM-W]
		\end{itemize}
		\STATE Update the model based on
		\begin{itemize}
			\item $\theta\leftarrow \theta - \eta \frac{\partial [(p^t_a/n)\ell(\Delta^t_{abc},1)+\gamma/2\|\Theta\|^2] }{\partial\theta }$,\\
			for PRFM, LFM-S, LFM-D
			\item $\theta\leftarrow \theta - \eta \frac{\partial [\Gamma(r(\x_{ab}))(p^t_a/n)\ell(\Delta^t_{abc},1)+\gamma/2\|\Theta\|^2] }{\partial\theta }$,\\
			for LFM-W
		\end{itemize}
		\ENDFOR
		\STATE {\bf Output}: the model $h^t$, or $\Theta^t=\{\w^t,V^t\}$
	\end{algorithmic}
\end{algorithm}

Then the learning process is to maximize some performance  which measures  the match between  $\pi(u_a,I_a,f)$ and $Y_u$, for all users $u_a$, $a=1,\ldots,n$. Specifically, we can use a general function $E[\pi(u_a,I_a,f),Y_a]$ to denote the ranking accuracy associated with each user and its item list $(u_a,I_a)$.  Then, the ranking accuracy in terms of a ranking metric, e.g., MAP, on the training data is re-written as below
\bqs
\frac{1}{n}\sum_{a=1}^n E[\pi(u_a,I_a,f),Y_a]\propto \sum E[\pi(u_a,I_a,f),Y_a].
\eqs
To maximize the ranking accuracy, we propose to minimize the following loss function:
\bqs
\arg\min_{f\in \mathcal{F}}\sum^n_{a=1}\{1-E[\pi(u_a,I_a,f),Y_a]\},
\eqs
where $\mathcal{F}$ is the set of all possible FM. Observation that this minimization is equivalent to maximizing the performance measures. However $E$ is a non-continuous function, it is difficult to optimize the loss function defined above. To solve this issue, we propose to minimize its upper bound as follows:
\bqs
\arg\min_{f\in \mathcal{F}}\sum^n_{a=1}\exp\{-E[\pi(u_a,I_a,f),Y_a]\}.
\eqs
The primary idea of applying boosting for Factorization Machine is to learn a set of component FMs and then create an ensemble of the components to predict the users' preferences on items. Specifically, we can use a linear combination of component FM as the final AdaFM model:
\bqs
f(\x)&=&\sum^T_{t=1}\alpha_t h^t(\x)
\eqs
where
\bqs
h^t=\w_t^\top\x+\sum^d_{l=1}\sum^d_{m=l+1}(V_tV_t^\top)_{lm}x_lx_m
\eqs
is the $t$-th component FM with small rank $k$ and $\alpha_t$ is a positive weight assigned to $h^t$ to determine its contribution in the final model. Therefor, for $f$ we can get an equivalent formulation as:
\bqs
f(\x)&=&\sum^T_{t=1}\alpha_t\Big[\w_t^\top\x+\sum^d_{l=1}\sum^d_{m=l+1}(V_tV_t^\top)_{lm}x_lx_m\Big]\\
&=&\bar{\w}_T^\top\x+\sum^d_{l=1}\sum^d_{m=l+1}(\bar{V}_T\bar{V}_T^\top)_{lm}x_lx_m
\eqs
where
\bqs
\bar{\w}_T=\sum^T_{t=1}\alpha_t \w_t\in\R^d,\quad \bar{V}_T=[\sqrt{\alpha_1}V_1,\ldots,\sqrt{\alpha_T}V_T]\in\R^{d\times kT}.
\eqs
This implies that the learnt $f$ is still a Factorization Machine, which rank $kT$.

In the training process, AdaFM runs for $T$ rounds, and one component FM is created at each round. At the $t$-th round, given the former $t-1$ components, the optimization problem is converted to
\bqs
(\alpha_t,h^t)=\arg\min_{(\alpha,h)}\sum^n_{a=1}\exp\{-E[\pi(u_a,I_a,f^{t-1}+\alpha h),Y_a]\}
\eqs
where $f^{t-1}=\sum^{t-1}_{s=1}\alpha_s h^s$.

To solve the above optimization, we first create an optimal component $h^t$ by using a re-weighting strategy, which assigns a dynamic weight $\beta^t_{a}$ for each user $u_a$. At each round, AdaFM increase the weights of the observed users for which their item lists are not ranked well by the ensemble components created so far. The learning process of the next component will then pay more attention to those "hard" users. Once, $h^t$ is given, the optimal $\alpha_t$ can be solved. Finally, the details of the AdaFM is summarized in Algorithm 1.
\setlength{\intextsep}{0pt}
\begin{algorithm}[ht]
\caption{The AdaFM Algorithm} \label{alg:AdaFM}
\begin{algorithmic}
\STATE {\bf Input}: Dataset $S=\{(u_a,I_a,Y_a)\}$, and performance measure $E$ and round $T$, and latent factor $k$
\STATE {\bf Initialize:} $p^1_a=1/n,\forall a$
\FOR{$t=1,2,\ldots,T$}
\STATE Solve $h^t=CA(S,\{p^t_a\},k)$
\STATE Compute $E[\pi(u_a,I_a,h^t),Y_a], \forall a$
\STATE Compute $\alpha_t=\frac{1}{2}\ln\frac{\sum_{a}p^t_a\{1+E[\pi(u_a,I_a,h^t),Y_a]\}}{\sum_{a}p^t_a\{1-E[\pi(u_a,I_a,h^t),Y_a]\}}$
\STATE Update $f^t=\sum^t_{s=1}\alpha_s h^s$
\STATE Compute $p^t_a=\frac{\exp\{-E[\pi(u_a,I_a,f^t),Y_a]\}}{\sum_{a}\exp\{-E[\pi(u_a,I_a,f^t),Y_a]\}}$ $\forall a$
\ENDFOR
\STATE {\bf Output}: the model $f=f^T$
\end{algorithmic}
\end{algorithm}

In algorithm 1, there is a key step using Component Algorithm (CA)
\bqs
h^t=CA(S,\{p^t_a\},k,E),
\eqs
for which the inputs are the data set $S$, the weights $\{p^t_a\}$, the latent factor $k$ and the performance measure $E$; and the output is a FMs model with latent factor $k$, which is obtained through maximizing
\bqs
\max_h\sum^n_{a=1}p^t_aE[\pi(u_a,I_a,h),Y_a].
\eqs
The specific algorithm to solve the above problem will be presented in the next subsection.

\subsubsection{Component Algorithm}
To construct component FMs, we can adopt the original FM, PRFM, or LambdaFM model. Specifically, for each user $u_a$ and an item  $i_{ab}\in I_a$, we can use the score of FM on $\x_{ab}$ to model the the relation between the user $u_a$ and item $i_{ab}$, as follows:
\bq
h_\Theta(\x) = \w^\top\x+ \sum^d_{l=1}\sum^d_{m=l+1}(\V\V^\top)_{lm} x_l x_m,
\eq
where $\w\in\R^d$ and $\V\in\R^{d\times k}$. At each round, the accuracy of the component $h^t$ can be evaluated by the ranking performance measure $E$ weighted by $p^t_a$. The optimal $h^t$ is then obtained by consistently optimizing the weighted ranking measure.

{\bf PRFM} is selected as the component algorithm to optimize AUC, which is chosen as the ranking metric. Given the weight distribution $p^t_a$, the accuracy of the component $h^t$ measured by weighted AUC, is defined as follows:
\bqs
wAUC&=&\sum_{a}\frac{p^t_a}{|H_a|}\sum_{H_a}\I(\pi^t_{ab}<\pi^t_{ac})\\
&=&\sum_{a}\frac{p^t_a}{|H_a|}\sum_{H_a}\I(h^t_{ab}>h^t_{ac})
\eqs
where $H_a=\{(b,c)|Y_{ab}\succ Y_{ac}\}$, $\pi^t_{ab}$ denotes the rank position of the item $i_{ab}$ in the list ranked by $h^t$ for $u_a$, and $h^t_{ab}=h^t(\x_{ab})$. Maximizing the weighted AUC is equivalent to minimizing the following loss function:
\bqs
\min_{h}\sum_{a}\frac{p^t_a}{|H_a|}\sum_{H_a}\I(h^t_{ab}\le h^t_{ac})
\eqs
To solve this problem, we replace the indicator function with a convex surrogate, i.e., the logistic regression loss function, as follows:
\bqs
\ell(\Delta^t_{abc},1)=\ln\left(1+\exp(-\Delta^t_{abc},1)\right)
\eqs
where $\Delta^t_{abc}=h^t_{ab}-h^t_{ac}$. The optimal component $h^t$ can be found by optimizing the following objective function:
\bqs
\min_{h}\frac{1}{n}\sum_{a}p^t_a\sum_{H_a}\frac{1}{|H_a|}\ell(\Delta^t_{abc},1)+\frac{\gamma}{2}\|\Theta\|^2
\eqs
where $\gamma>0$ is a regularization parameter. The problem above can be solved by stochastic gradient descent, which firstly uniformly sample one user $u_a$ from all the users, then sample a pair $(b,c)$ from $H_a$, and finally update the model based on the following method:
\bqs
\theta&\leftarrow& \theta - \eta \frac{\partial [(p^t_a/n)\ell(\Delta^t_{abc},1)+\gamma/2\|\Theta\|^2] }{\partial\theta }
\eqs
where $\theta\in \{w_l, v_{l,m}\}$, and $\eta>0$ is the learning rate. To calculate the gradient of the objective with respect to $\theta$,  we can firstly derive the gradient using the property of Multi-linearity:
\begin{displaymath}
\frac{\partial h^t_{ab} }{\partial \theta } = \left\{ \begin{array}{ll}
x_{ab}^l & \textrm{if $\theta$ is $w_l$}\\
x^l_{ab}\sum^d_{r=1}v_{r,m}x^r_{ab}-v_{l,m}(x^l_{ab})^2 & \textrm{if $\theta$ is $v_{l,m}$}\\
\end{array} \right.
\end{displaymath}
Then, if we denote
\bqs
\lambda_{abc}=\frac{\partial \ell(\Delta^t_{abc},1) }{\Delta^t_{abc}}=\frac{-\exp(-\Delta^t_{abc})}{1+\exp(-\Delta^t_{abc})}
\eqs
the stochastic gradient for $w_l$ can be computed as
\bqs
&&\hspace{-0.2in}\frac{\partial [(p^t_a/n)\ell(\Delta^t_{abc},1)+\gamma/2\|\Theta\|^2] }{\partial w_l }=\frac{p^t_a}{n}\frac{\partial \ell(\Delta^t_{abc},1) }{\Delta^t_{abc}}\frac{\partial\Delta^t_{abc}}{\partial w_l}+\gamma w_l\\
&&=(p^t_a/n)\lambda_{abc}(x^l_{ab}-x^l_{ac})+\gamma w_l
\eqs
and the stochastic gradient for $v_{l,m}$ can be computed as
\bqs
&&\hspace{-0.2in}\frac{\partial [(p^t_a/n)\ell(\Delta^t_{abc},1)+\gamma/2\|\Theta\|^2] }{\partial v_{l,m} }\\
&&\hspace{-0.2in}=\frac{p^t_a}{n}\frac{\partial \ell(\Delta^t_{abc},1) }{\Delta^t_{abc}}\frac{\partial\Delta^t_{abc}}{\partial v_{l,m}}+\gamma v_{l,m}\\
&&\hspace{-0.2in}=\frac{p^t_a}{n}\lambda_{abc}\left\{\sum^d_{r=1}v_{r,m}(x_{ab}^r x_{ab}^l-x_{ac}^r x_{ac}^l)-v_{l,m}[(x_{ab}^l)^2-(x_{ac}^l)^2 )]\right\}\\
&&+\gamma v_{l,m}.
\eqs

{\bf LambdaFM} is selected to optimize NDCG, which is chosen as the performance metric. For this case, we can adopt the lambda sampling strategies~\cite{DBLP:conf/cikm/YuanGJCYZ16} instead of the uniform sampling one, i.e, the popularity based Static Sampler~\eqref{ss}, Rank-Aware Dynamic Sampler~\eqref{cs}, and Rank-aware Weighted Approximation~\eqref{rs}.

Finally, the algorithm for building the component is summarized in the following Algorithm 2.

\setlength{\intextsep}{0pt}
\begin{figure*}
    \centering
    \subfigure[Yahoo]{
    \centering
    \label{fig:yahoo-fm-auc} 
    \includegraphics[width=50mm]{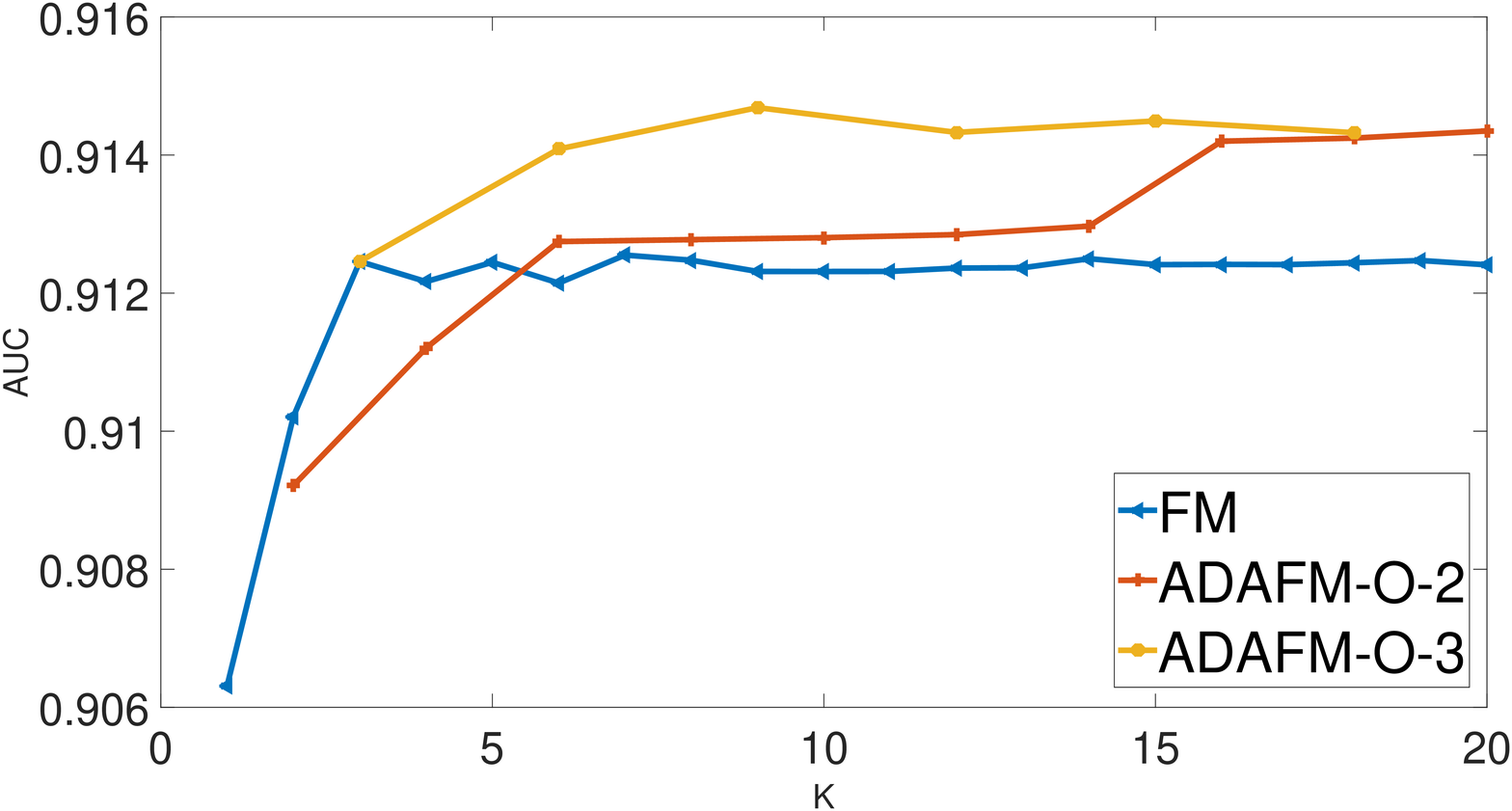}}
    \subfigure[Yelp]{
    \centering
    \label{fig:yelp-prfm-auc} 
    \includegraphics[width=50mm]{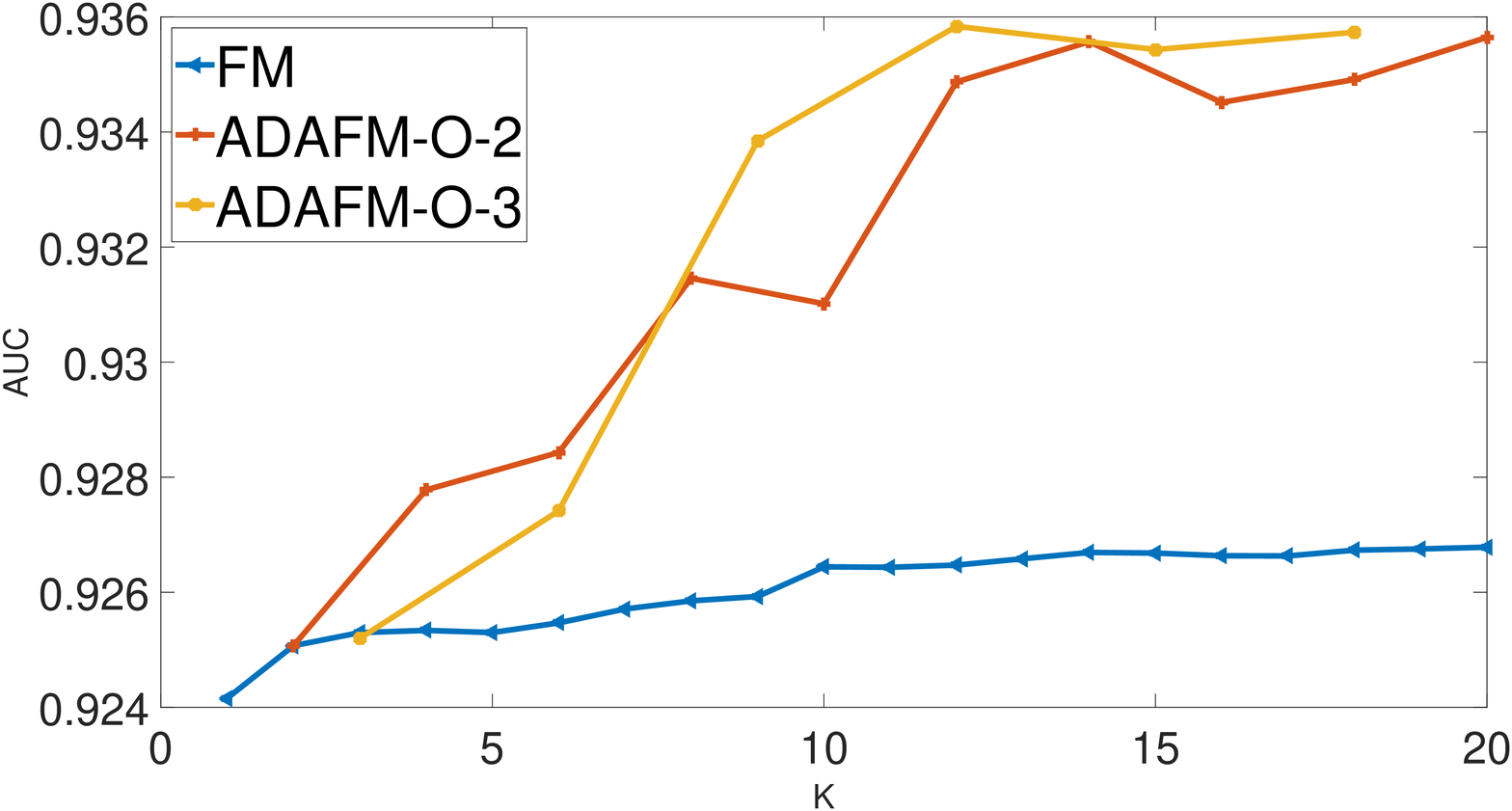}}
     \subfigure[Lastfm]{
    \centering
    \label{fig:lastfm-fm-auc} 
    \includegraphics[width=50mm]{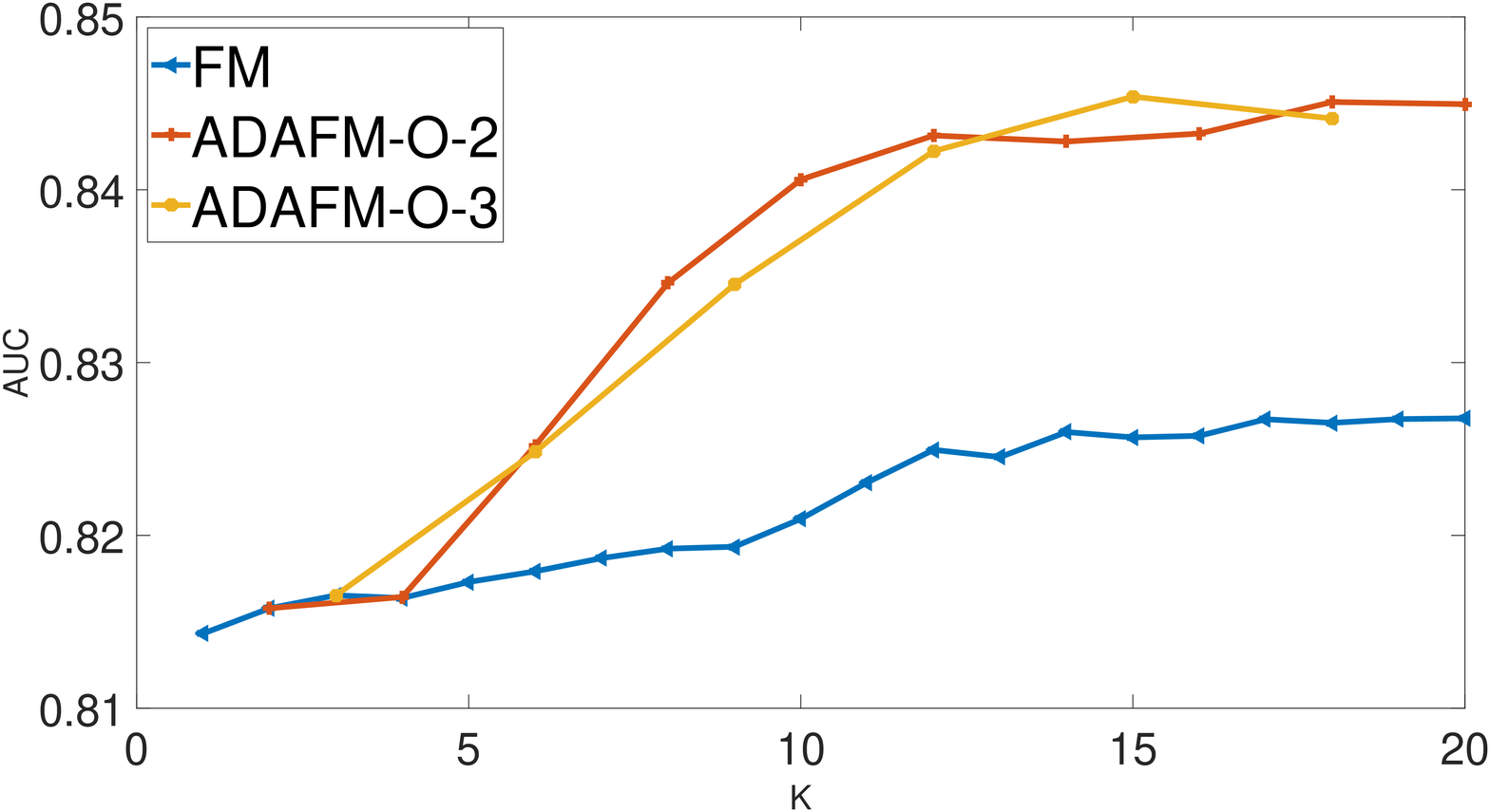}}
    \caption{FM based learner's results on different datasets. For FM, the horizontal axis denotes the latent dimension. For AdaFM-O, the horizontal axis denotes the weak learner's latent dimension multiplied by the number of weak learners.}
    \label{fig:fm-auc}
\end{figure*}

\setlength{\intextsep}{0pt}
\begin{figure*}
	\centering
	\subfigure[Yahoo]{
		\centering
		\label{fig:yahoo-prfm-auc} 
		\includegraphics[width=50mm]{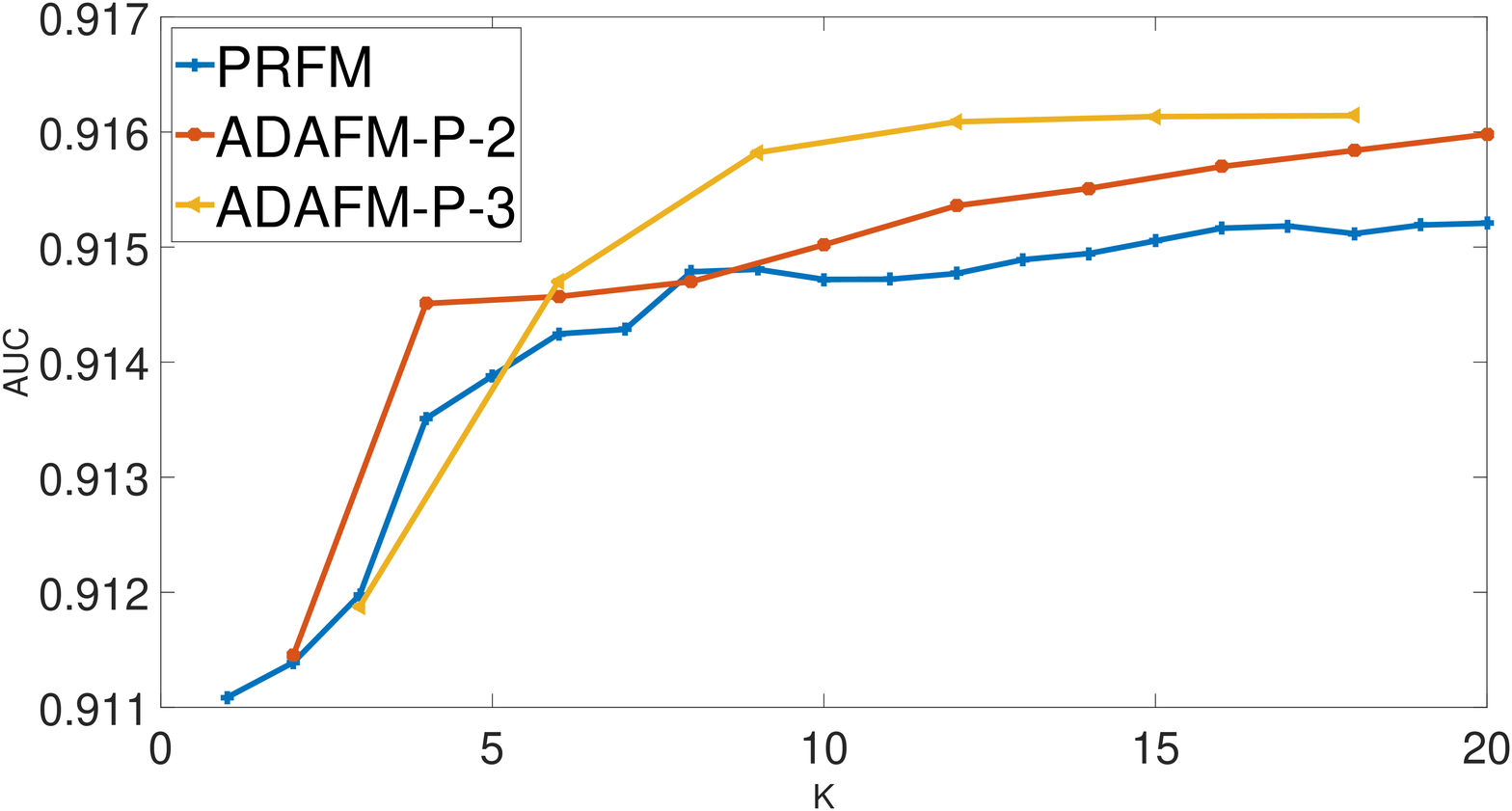}}
	\subfigure[Yelp]{
		\centering
		\label{fig:yelp-prfm-auc} 
		\includegraphics[width=50mm]{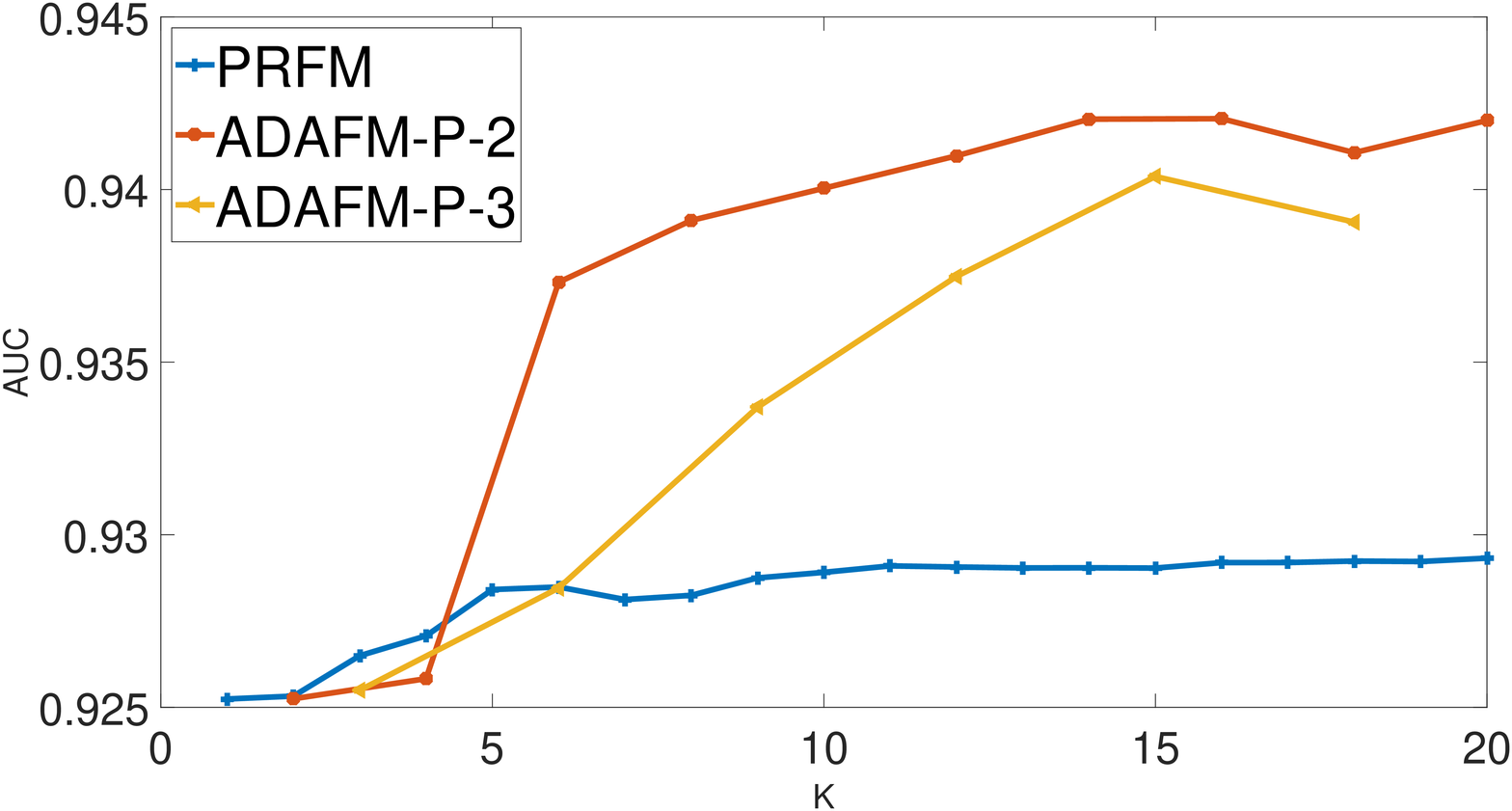}}
	\subfigure[Lastfm]{
		\centering
		\label{fig:lastfm-prfm-auc} 
		\includegraphics[width=50mm]{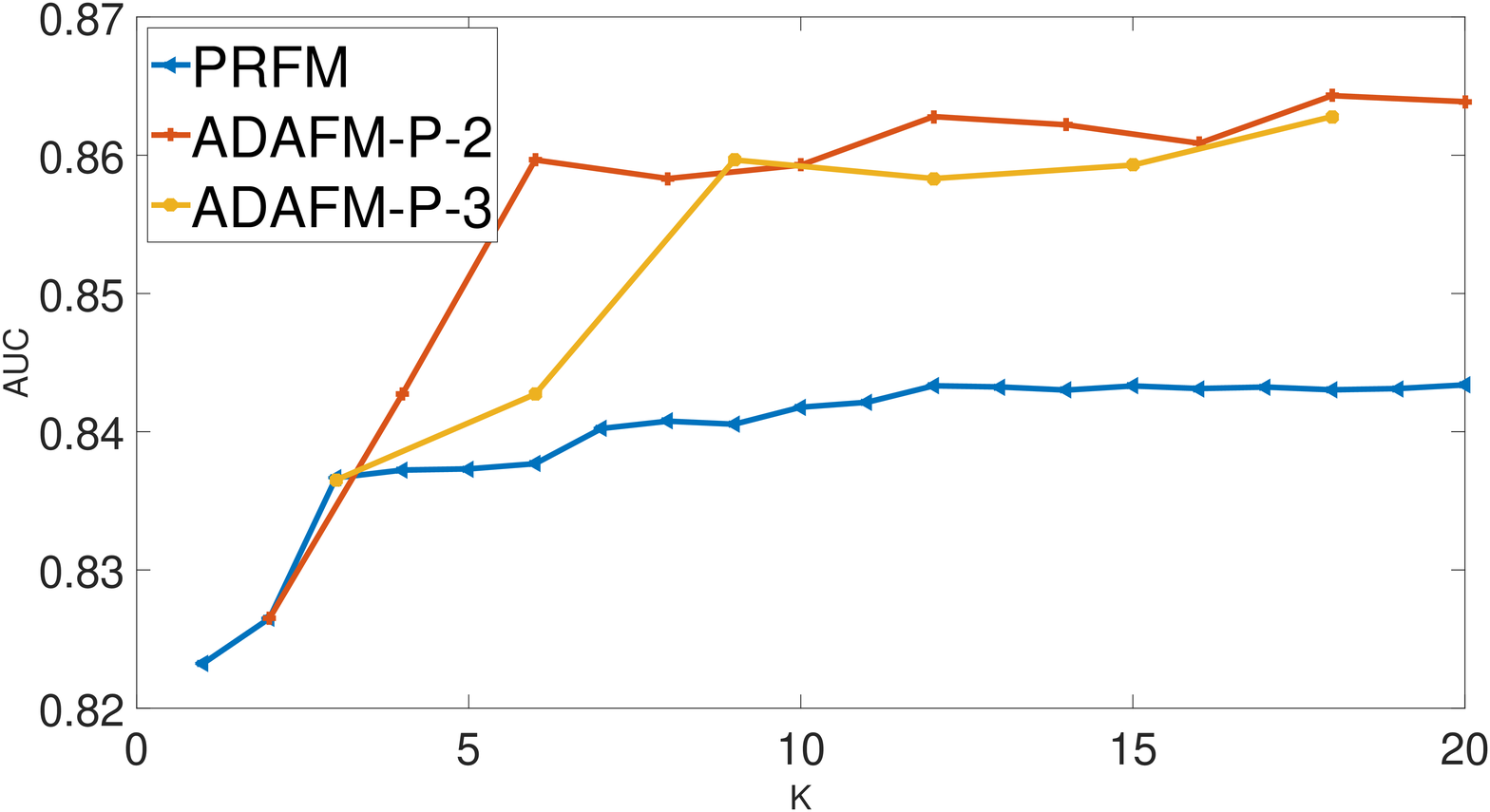}}
	\caption{PRFM based learner's results on different datasets. The horizontal axis has the same meansing as Figure ~\ref{fig:fm-auc}. }
	\label{fig:prfm-auc}
\end{figure*}

\section{Experiments}
In this section, we report a comprehensive suite of experimental results that help evaluate the  performance of our proposed AdaFM algorithm on several recommendation tasks.  The experiments are designed to answer the following open questions:
(1) Whether the proposed boosting approach is effective to improve the ranking performances significantly? (2) Whether the weak learner's latent dimension has a great effect on ranking performances.

\subsection{Experimental Testbed}
\begin{table}[hptp!]
	\centering
	\caption{Basic statistics of datasets. Each entry indicates whether a user has interacted with an item.}
	\label{tab:data}
	\begin{tabular}{lrrrrr}
		\toprule
		Datasets&\#Users & \#Items & \#Entries\\
		\midrule
		Yelp & 17,526 & 85,539 & 875,955\\
		Lastfm& 992& 60,000 & 759,391\\
		Yahoo & 2,450& 6,518 & 107,334\\
		\bottomrule
	\end{tabular}
\end{table}

We evaluate our proposed algorithm against several baselines on three publicly available Collaborative Filtering (CF) datasets, i.e., Yelp\footnote{\url{https://www.yelp.com/dataset\_challenge}} (user-venue pairs), Lastfm\footnote{\url{http://www.dtic.upf.edu/~ocelma/MusicRecommendationDataset/lastfm-1K.html}} (user-music pairs), and Yahoo music\footnote{\url{https://webscope.sandbox.yahoo.com/catalog.php?datatype=r}} (user-music pairs).
To speed up the experiments, we perform the following sampling strategies on these datasets.
For Yelp, we filter out the users with less than 20 interactions.
For Yahoo, we derive a smaller dataset by randomly sampling a subset of users and items from the original dataset.
The statistics of the datasets after preprocessing are summarized in Table \ref{tab:data}.

To test the performances of our proposed AdaFM framework under different optimization targets, we adopt two standard ranking metrics: Area Under ROC Curve (AUC) and Normalized Discounted Cumulative Gain (NDCG).


\begin{table*}
	\caption{Performance on NDCG. The best result is indicated in bold.}
	\centering
	\label{tab:ndcg-optim}
	\begin{tabular}{lrrrrrrrrr}
		\toprule
		Datasets & \textbf{FM} & \textbf{PRFM} & \textbf{LFM-S} & \textbf{AdaFM-S} & \textbf{LFM-D} & \textbf{AdaFM-D} & \textbf{LFM-W} & \textbf{AdaFM-W}\\
		\midrule
		Yelp & 0.204 & 0.205 & 0.217 & 0.225 & 0.215 & \textbf{0.228}   & 0.221 & 0.227\\
		
		Yahoo  &0.382 & 0.383 & 0.386 & 0.407 & 0.392 & 0.408 & 0.395 & \textbf{0.410} \\
		
		\bottomrule
	\end{tabular}
\end{table*}

\subsection{Comparison Algorithms}
Our proposed AdaFM is a general framework for improving the performances of FM derived algorithms.
Thus, we compare the performances of the following FM derived algorithms and their corresponding enhanced models using our proposed AdaFM framework.
\begin{itemize}[leftmargin=*] \setlength{\itemsep}{-\itemsep}
    \item The \textbf{O}riginal FM that is designed for the rating prediction task, and its enhanced model using AdaFM and we name it AdaFM-O for short;
    \item \textbf{P}ariwise \textbf{R}anking FM (PRFM), which aims to maximize the AUC metric, and its adaptive version (AdaFM-P);
    \item LambdaFM, which is designed to maximize the NDCG metric. We use three different sampling strategies to form the list pairs, i.e., \textbf{S}tatic sampler, \textbf{D}ynamic sampler, and rank-a\textbf{w}are sampler, as described in Section ~\ref{sec:lfm}, and we name them as LFM-S, LFM-D, and LFM-W, respectively. We also name their adaptive versions as AdaFM-S, AdaFM-D, and AdaFM-W, respectively.
\end{itemize}




\begin{figure*}
	\centering
	\subfigure[AdaFM-S]{
		\label{fig:yahoo-lfms-ndcg} 
		\includegraphics[width=50mm]{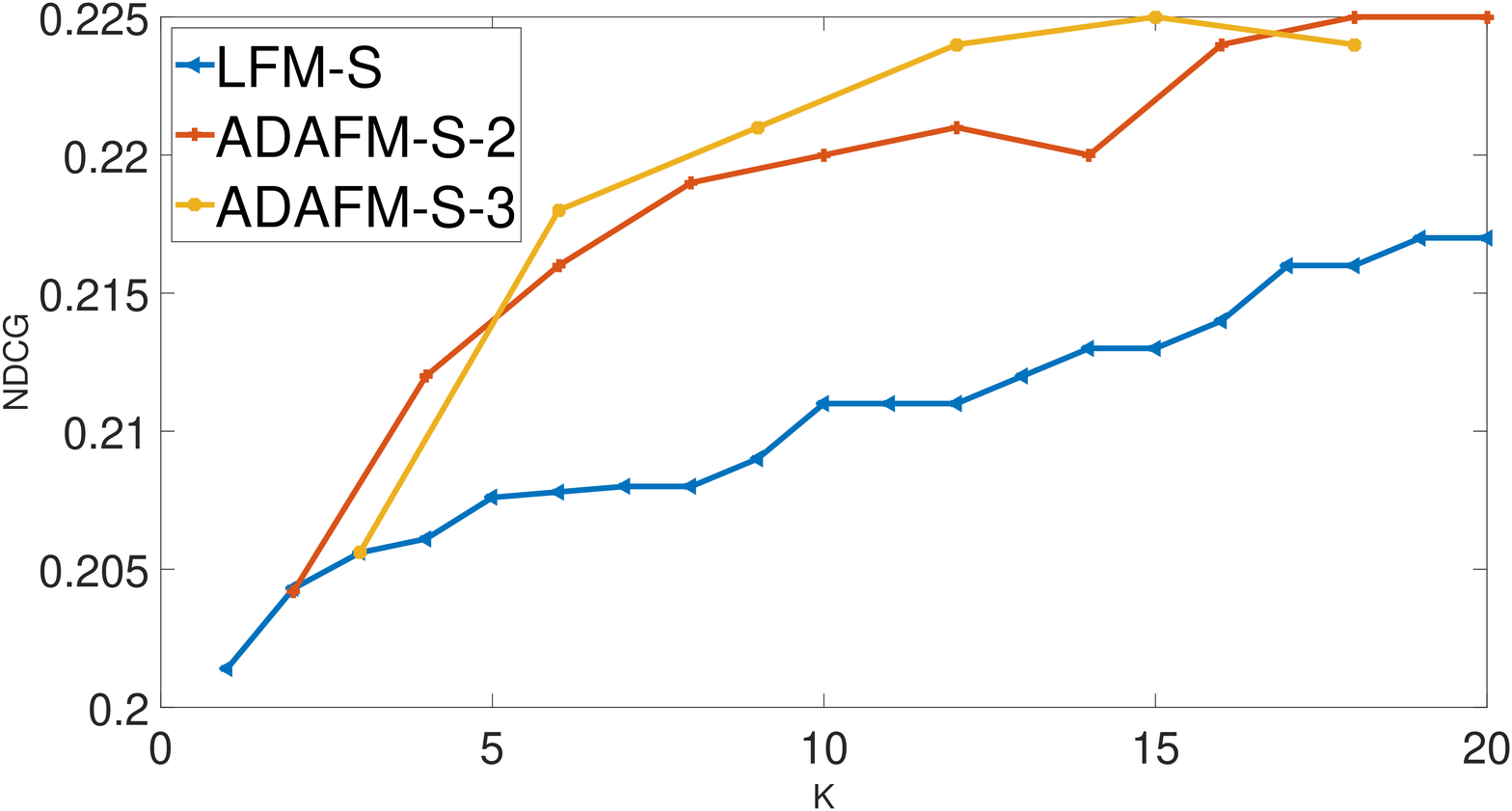}
	}
	\subfigure[AdaFM-D]{
		\label{fig:yahoo-lfmd-ndcg} 
		\includegraphics[width=50mm]{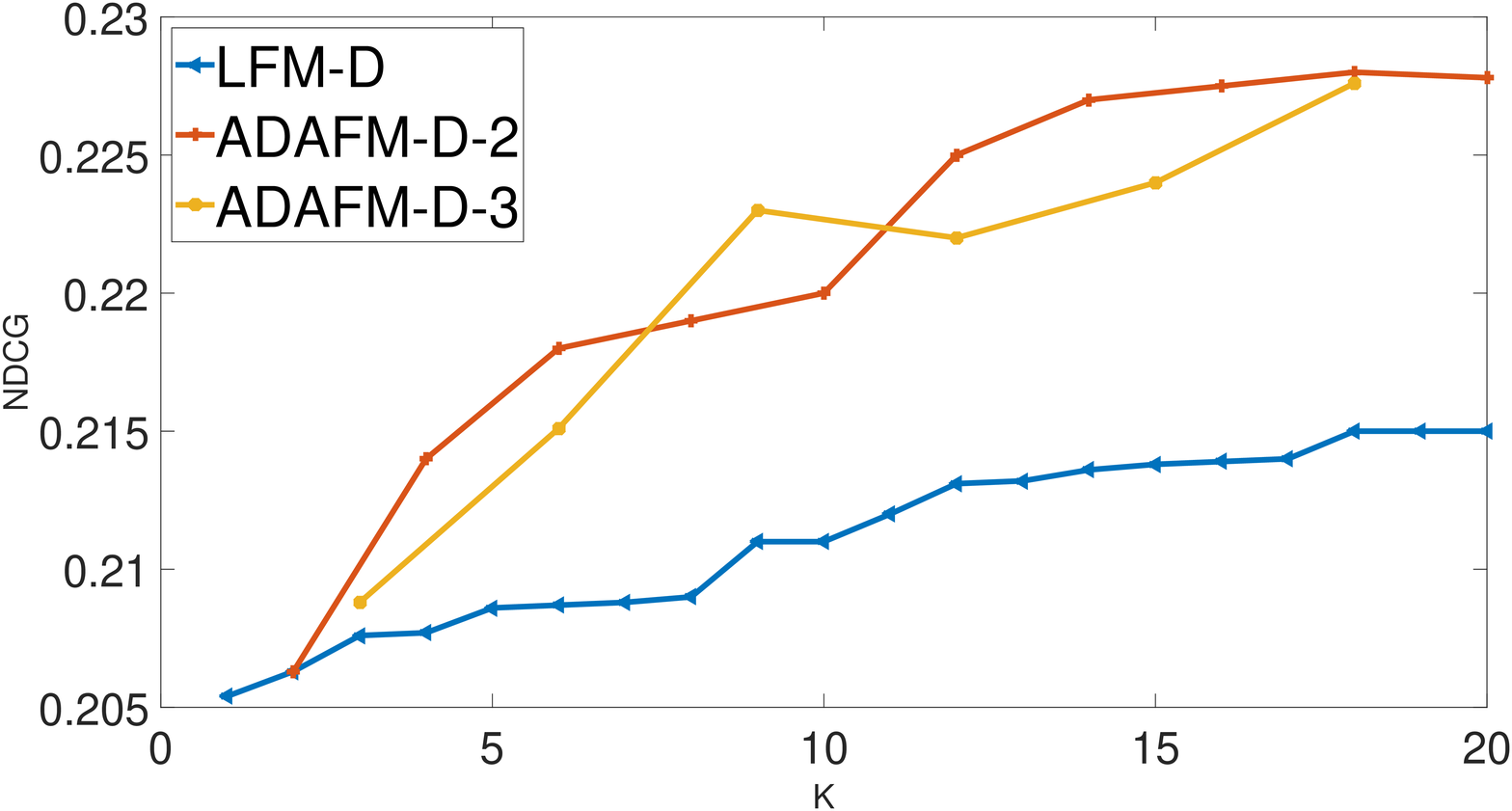}
	}
	\subfigure[AdaFM-W]{
		\label{fig:yahoo-lfmd-ndcg} 
		\includegraphics[width=50mm]{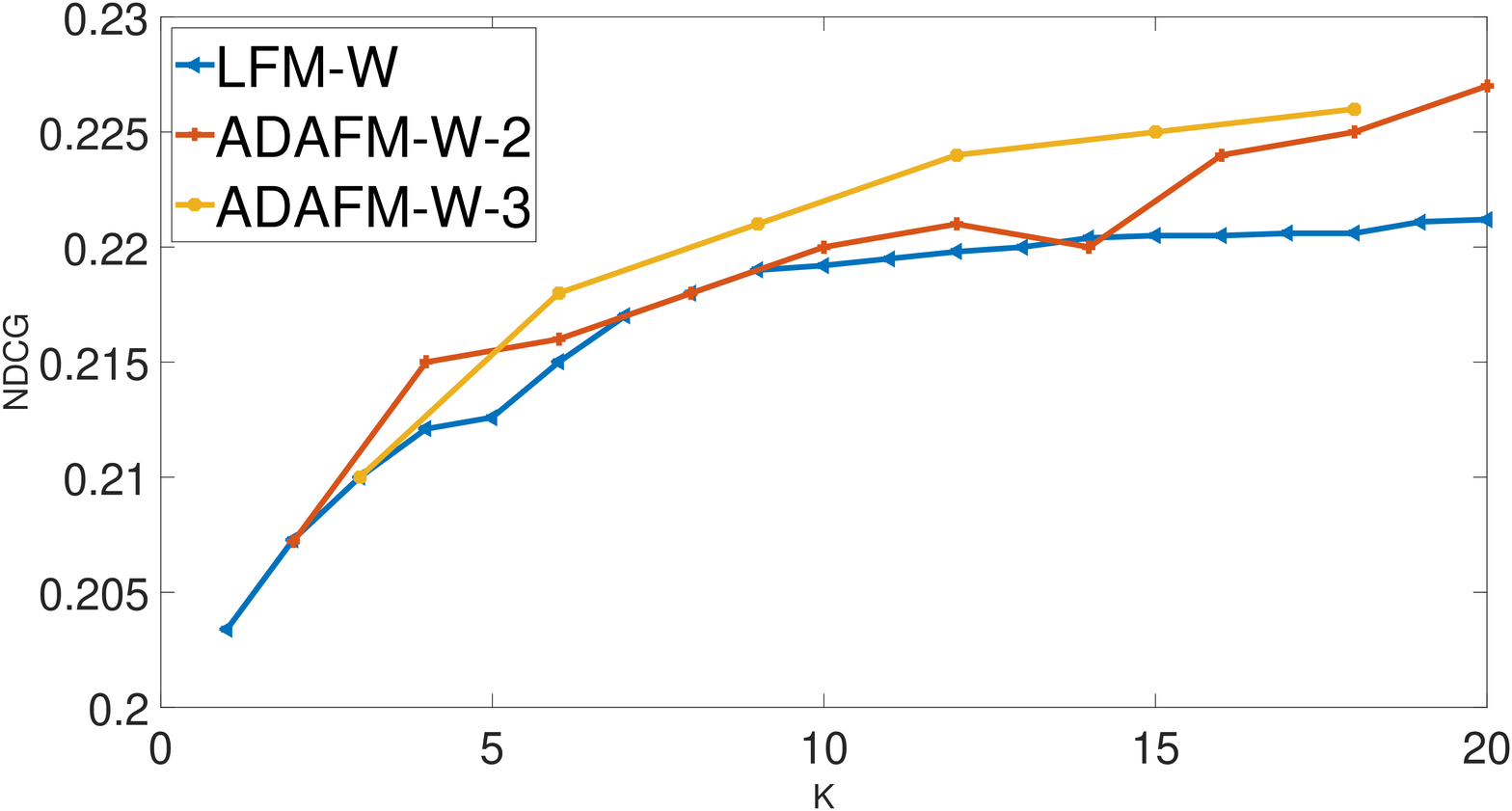}
	}
	\caption{NDCG results on Yahoo dataset. The horizontal axis has the same meansing as Figure ~\ref{fig:fm-auc}. }
	\label{fig:yahoo-lfm} 
\end{figure*}

\begin{figure*}
	\centering
	\subfigure[AdaFM-S]{
		\label{fig:yelp-lfms-ndcg} 
		\includegraphics[width=50mm]{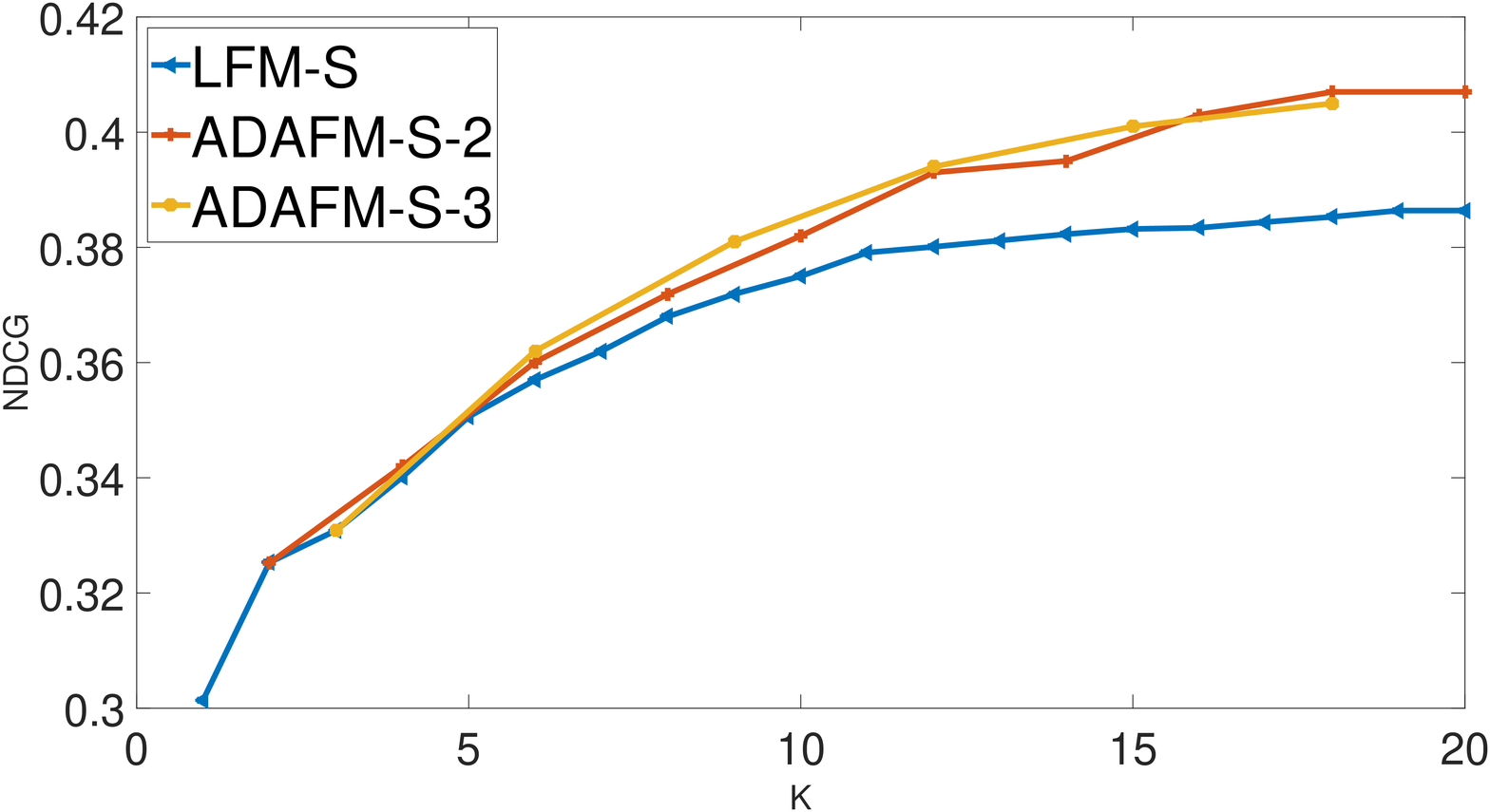}
	}
	\subfigure[AdaFM-D]{
		\label{fig:yelp-lfmd-ndcg} 
		\includegraphics[width=50mm]{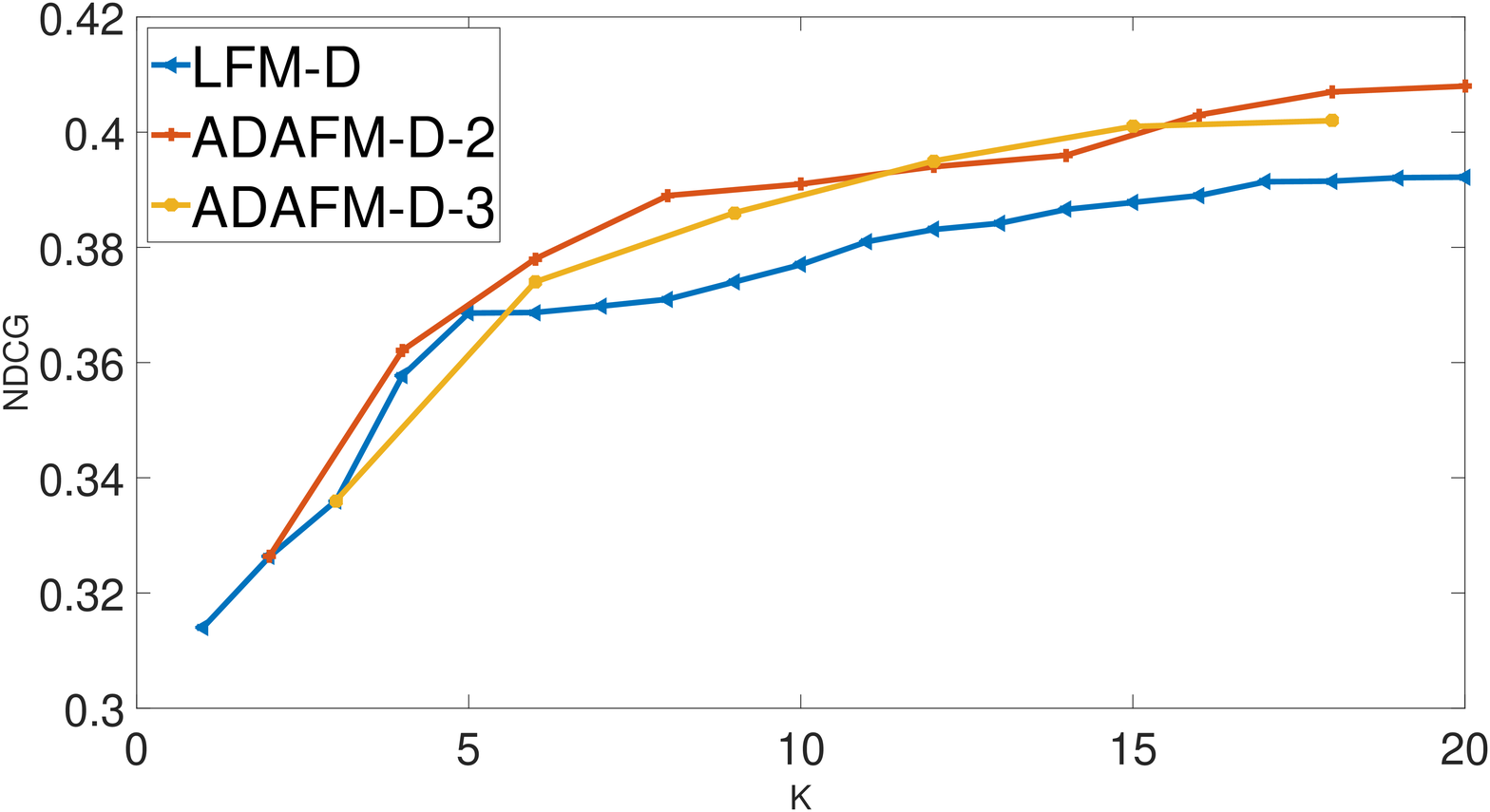}
	}
	\subfigure[AdaFM-W]{
		\label{fig:yelp-lfmd-ndcg} 
		\includegraphics[width=50mm]{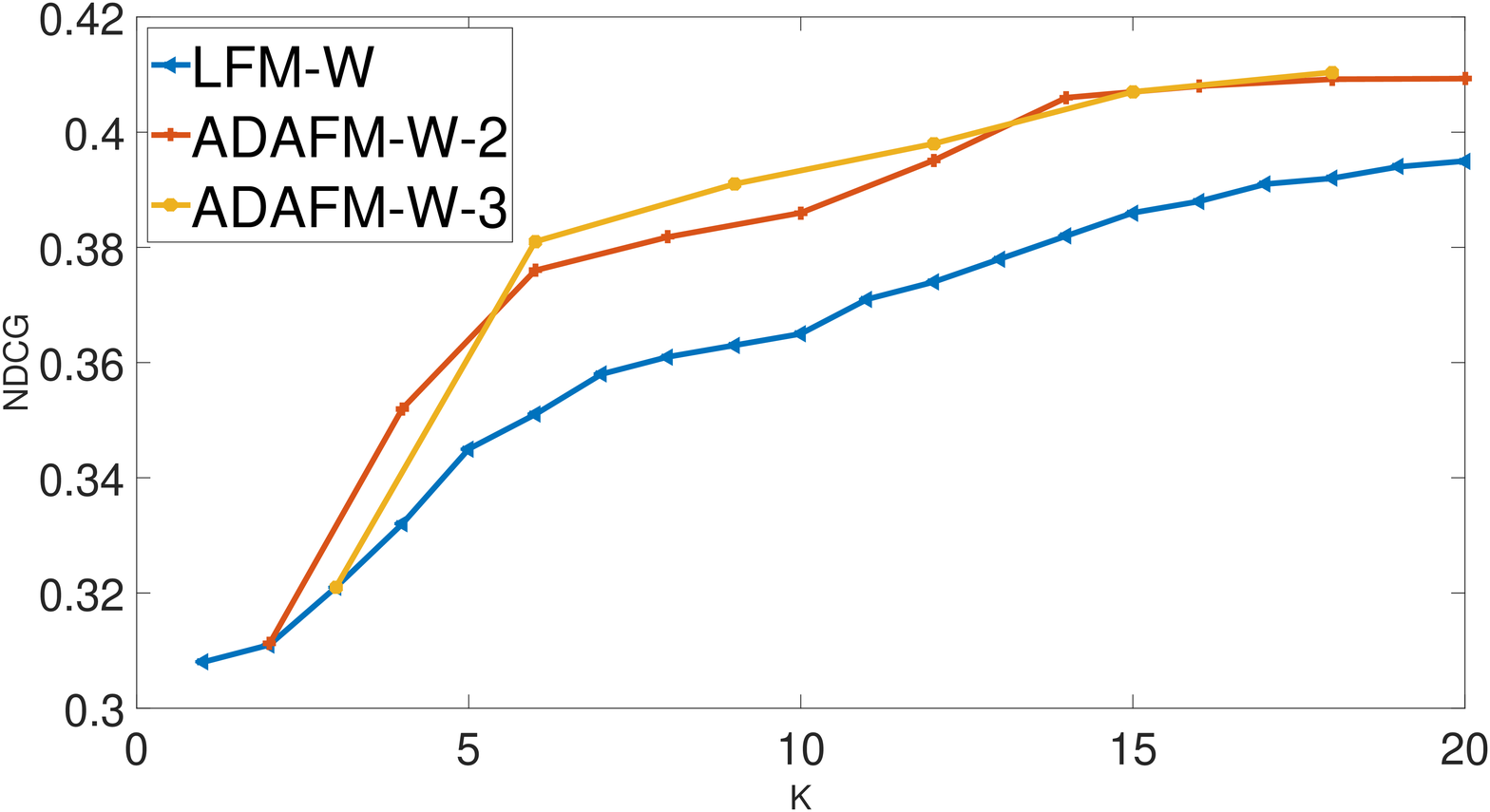}
	}
	\caption{NDCG results on Yelp dataset. The horizontal axis has the same meansing as Figure ~\ref{fig:fm-auc}. }
	\label{fig:yelp-lfm} 
\end{figure*}

\subsection{Hyper-parameter Settings}
The main parameters to be tuned in our experiments are as follows:

\noindent  \textbf{Learning rate $\eta$} : For base learners, we first apply the 5-fold cross validation to find the best $\eta$ for FM when $k=2$, and then use the same $\eta$ for the PRFM, LambdaFM, AdaFM.

\noindent \textbf{Latent dimension $k$}: In order to compare the performance of AdaFM and the base learners, we simply choose the latent dimension of AdaFM from $k \in \{2,3\}$, and range the latent dimension of the FM derived algorithms in $k \in [1, 20]$.



\noindent  \textbf{Regularization $\gamma$}: FM derived algorithms have several regularization parameters, including $\gamma_{w_l}$ and $\gamma_{v_{l, m}}$, which represent the regularization parameters of $w_l$ and $v_{l,m}$ , respectively.
During the experiments, we select the best values of $\gamma$ in $ \{0.5, 0.1, 0.05, 0.01, 0.005 \}$ for each FM derived algorithm.
For simplicity, in our experiments, we restrict  $\gamma_{w_{l}}$ and $\gamma_{v_{l,m}}$ to have the same value of $\gamma$.

\noindent  \textbf{Distribution coefficient $\rho$}: $\rho$ controls the sampling probability of Lambda FM, and is usually affected by data distribution.
Thus, we select the best values of $\rho$ for LFM-S, LFM-D, and LFM-W in $(0, 1]$.

\begin{table}
  \caption{Performance comparision on AUC. The best result is indicated in bold.}
  \label{tab:auc-optim}
  \small
  \begin{tabular}{lrrrrr}
  	
    \toprule
    Datasets&  \textbf{FM} & \textbf{AdaFM-O} & \textbf{PRFM} & \textbf{AdaFM-P}  \\
    \midrule
      Yelp &  0.911 & 0.914 & 0.915 & \textbf{0.916} \\
      Lastfm & 0.826 & 0.845 & 0.843 & \textbf{0.864}\\
      Yahoo & 0.925 & 0.936 & 0.929 & \textbf{0.942} \\
  \bottomrule
\end{tabular}
\vspace{5pt}
\end{table}


\subsection{Performance Evaluation}

\subsubsection{AUC Optimization}
We start by evaluating the effectiveness of our proposed AdaFM framework on AUC maximization task.
The detailed results are presented in Figure \ref{fig:fm-auc} and  \ref{fig:prfm-auc}, and Table \ref{tab:auc-optim}. Several insightful observations can be made.

First, after combine the Adaptive Boosting and FM, the final results are 
increased. As shown in Figure \ref{fig:fm-auc}, when use FM as weak learner, compare with the FM, on Lastfm dataset, we get an  2.32\% improvement.
And as shown in Figure \ref{fig:prfm-auc}, when use PRFM as weak learner, compare with the PRFM, on Lastfm, we get an 2.49\% improvement.


Second, AdaFM shows better results by using less parameters. This is clearly evident in Figure \ref{fig:fm-auc} and \ref{fig:prfm-auc}. For example, in all datasets, AdaFM with four weak learners (which latent dimension is 2) achieves a comparable or even better results than the base FM and PRFM with $k=20$. The results are encouraging as it shows in the cases when the base FM and PRFM stuck in a certain local optimum, our proposed boosting framework can help to achieve better results.

Last but not least, as shown in Table \ref{tab:auc-optim}, the AdaFM-P has the best results on all the datasets. This shows when using a better weak learner, i.e., PRFM in our case, the AdaFM method achieves better results. This further demonstrates the effectiveness of our boosting framework.

%

\subsubsection{NDCG Optimization}
We proceed to evaluate the effectiveness of our AdaFM framework on NDCG maximization task.
We use LambdaFM with different samplers as our baselines, which are designed to optimize the NDCG metric. More specifically,
we consider three variants of LambdaFM, i.e., LFM-S, LFM-D, and LFM-W, with their corresponding boosted versions, i.e., AdaFM-S, AdaFM-D and AdaFM-W.
As shown in the Table \ref{tab:ndcg-optim}, LambdaFM is better than FM and PRFM, as LambdaFM is designed to optimize the NDCG metric. But our AdaFM methods outperform all the three variants of LambdaFM: LFM-S, LFM-D, and LFM-W.
Specifically, on Yelp dataset, comparing with the original algorithm, AdaFM-S, AdaFM-D, and AdaFM-W get 3.6\%, 6.04\% and 2.7\% improvement, respectively.
On Yahoo dataset, AdaFM-S, AdaFM-D, and AdaFM-W get 4.8\%, 2.19\% and 3.8\% improvement, respectively.

And as shown in Figure \ref{fig:yahoo-lfm} and \ref{fig:yelp-lfm}, the findings are similar to Figure \ref{fig:fm-auc} and \ref{fig:prfm-auc} where our proposed AdaFM achieves better results.

\subsection{Effect of Latent Dimension}


In this section, we study whether weak learner's latent dimension affect the final results of our proposed AdaFM.

From the experiments  in Figure~\ref{fig:fm-auc}, ~\ref{fig:prfm-auc}, ~\ref{fig:yahoo-lfm}, and ~\ref{fig:yelp-lfm}, we find that: (1) with the increase of weak learner numbers, the performances of our proposed AdaFM first increase and then become stable, no matter the latent dimension of weak learners; (2) AdaFM tends to have similar performance even when the latent dimension of the weak learners are different.
For example, the AUC performances of AdaFM-O-2 and AdaFM-O-3 both increase with the weak learner nubmers on Lastfm (i.e., Figure ~\ref{fig:lastfm-fm-auc}), however, they achieve quite similar AUC performance after a certain weak learner numbers (i.e., 0.845 vs. 0.844).
This finding indicates that
it is easy for our proposed AdaFM to tune model parameters in practice.


\section{Conclusions}

In this paper, we first proposed a novel Adaptive Boosting framework of factorization machine(AdaFM), which combines the advantages of adaptive boosting and FM.
Our proposed AdaFM is a general framework that can be used to improve the performance of all the existing FM derived algorithms, e.g., FM, PRFM, and LambdaFM.
We then presented the details of how to combine adaptive boosting technique and FM derived models.
We finally performed thorough experiments to evaluate our model performance on three real public datasets.
The results demonstrated that AdaFM is able to improve the prediction performances in both AUC and NDCG maximization tasks.


\bibliographystyle{siam}
\bibliography{sigproc}


\begin{thebibliography}{00}


\ifx \showCODEN    \undefined \def \showCODEN     #1{\unskip}     \fi
\ifx \showDOI      \undefined \def \showDOI       #1{{\tt DOI:}\penalty0{#1}\ }
  \fi
\ifx \showISBNx    \undefined \def \showISBNx     #1{\unskip}     \fi
\ifx \showISBNxiii \undefined \def \showISBNxiii  #1{\unskip}     \fi
\ifx \showISSN     \undefined \def \showISSN      #1{\unskip}     \fi
\ifx \showLCCN     \undefined \def \showLCCN      #1{\unskip}     \fi
\ifx \shownote     \undefined \def \shownote      #1{#1}          \fi
\ifx \showarticletitle \undefined \def \showarticletitle #1{#1}   \fi
\ifx \showURL      \undefined \def \showURL       #1{#1}          \fi
\providecommand\bibfield[2]{#2}
\providecommand\bibinfo[2]{#2}
\providecommand\natexlab[1]{#1}
\providecommand\showeprint[2][]{arXiv:#2}

\bibitem[\protect\citeauthoryear{Burges, Ragno, and Le}{Burges
  et~al\mbox{.}}{2006}]%
        {DBLP:conf/nips/BurgesRL06}
\bibfield{author}{\bibinfo{person}{Christopher J.~C. Burges},
  \bibinfo{person}{Robert Ragno}, {and} \bibinfo{person}{Quoc~Viet Le}.}
  \bibinfo{year}{2006}\natexlab{}.
\newblock \showarticletitle{Learning to Rank with Nonsmooth Cost Functions}. In
  \bibinfo{booktitle}{{\em Advances in Neural Information Processing Systems
  19, Proceedings of the Twentieth Annual Conference on Neural Information
  Processing Systems, Vancouver, British Columbia, Canada, December 4-7,
  2006}}. \bibinfo{pages}{193--200}.
\newblock
\showURL{%
\url{http://papers.nips.cc/paper/2971-learning-to-rank-with-nonsmooth-cost-functions}}


\bibitem[\protect\citeauthoryear{Cheng, Xia, Zhang, King, and Lyu}{Cheng
  et~al\mbox{.}}{2014}]%
        {cheng2014gradient}
\bibfield{author}{\bibinfo{person}{Chen Cheng}, \bibinfo{person}{Fen Xia},
  \bibinfo{person}{Tong Zhang}, \bibinfo{person}{Irwin King}, {and}
  \bibinfo{person}{Michael~R. Lyu}.} \bibinfo{year}{2014}\natexlab{}.
\newblock \showarticletitle{Gradient boosting factorization machines}. In
  \bibinfo{booktitle}{{\em RecSys'14}}. ACM, \bibinfo{pages}{265--272}.
\newblock


\bibitem[\protect\citeauthoryear{Cortes and Mohri}{Cortes and Mohri}{2003}]%
        {DBLP:conf/nips/CortesM03}
\bibfield{author}{\bibinfo{person}{Corinna Cortes} {and}
  \bibinfo{person}{Mehryar Mohri}.} \bibinfo{year}{2003}\natexlab{}.
\newblock \showarticletitle{{AUC} Optimization vs. Error Rate Minimization}. In
  \bibinfo{booktitle}{{\em Advances in Neural Information Processing Systems 16
  [Neural Information Processing Systems, {NIPS} 2003, December 8-13, 2003,
  Vancouver and Whistler, British Columbia, Canada]}}.
  \bibinfo{pages}{313--320}.
\newblock
\showURL{%
\url{http://papers.nips.cc/paper/2518-auc-optimization-vs-error-rate-minimization}}


\bibitem[\protect\citeauthoryear{Cremonesi, Koren, and Turrin}{Cremonesi
  et~al\mbox{.}}{2010}]%
        {DBLP:conf/recsys/CremonesiKT10}
\bibfield{author}{\bibinfo{person}{Paolo Cremonesi}, \bibinfo{person}{Yehuda
  Koren}, {and} \bibinfo{person}{Roberto Turrin}.}
  \bibinfo{year}{2010}\natexlab{}.
\newblock \showarticletitle{Performance of recommender algorithms on top-n
  recommendation tasks}. In \bibinfo{booktitle}{{\em Proceedings of the 2010
  {ACM} Conference on Recommender Systems, RecSys 2010, Barcelona, Spain,
  September 26-30, 2010}}. \bibinfo{pages}{39--46}.
\newblock
\showDOI{%
\url{http://dx.doi.org/10.1145/1864708.1864721}}


\bibitem[\protect\citeauthoryear{Cristianini and Shawe{-}Taylor}{Cristianini
  and Shawe{-}Taylor}{2010}]%
        {DBLP:books/daglib/0026018}
\bibfield{author}{\bibinfo{person}{Nello Cristianini} {and}
  \bibinfo{person}{John Shawe{-}Taylor}.} \bibinfo{year}{2010}\natexlab{}.
\newblock \bibinfo{booktitle}{{\em An Introduction to Support Vector Machines
  and Other Kernel-based Learning Methods}}.
\newblock \bibinfo{publisher}{Cambridge University Press}.
\newblock
\showISBNx{978-0-521-78019-3}


\bibitem[\protect\citeauthoryear{Duffy and Helmbold}{Duffy and
  Helmbold}{2002}]%
        {duffy2002boosting}
\bibfield{author}{\bibinfo{person}{Nigel Duffy} {and} \bibinfo{person}{David
  Helmbold}.} \bibinfo{year}{2002}\natexlab{}.
\newblock \showarticletitle{Boosting methods for regression}.
\newblock \bibinfo{journal}{{\em Machine Learning\/}} \bibinfo{volume}{47},
  \bibinfo{number}{2-3} (\bibinfo{year}{2002}), \bibinfo{pages}{153--200}.
\newblock


\bibitem[\protect\citeauthoryear{Freund, Iyer, Schapire, and Singer}{Freund
  et~al\mbox{.}}{2003}]%
        {DBLP:journals/jmlr/FreundISS03}
\bibfield{author}{\bibinfo{person}{Yoav Freund}, \bibinfo{person}{Raj~D. Iyer},
  \bibinfo{person}{Robert~E. Schapire}, {and} \bibinfo{person}{Yoram Singer}.}
  \bibinfo{year}{2003}\natexlab{}.
\newblock \showarticletitle{An Efficient Boosting Algorithm for Combining
  Preferences}.
\newblock \bibinfo{journal}{{\em Journal of Machine Learning Research\/}}
  \bibinfo{volume}{4} (\bibinfo{year}{2003}), \bibinfo{pages}{933--969}.
\newblock
\showURL{%
\url{http://www.jmlr.org/papers/v4/freund03a.html}}


\bibitem[\protect\citeauthoryear{Freund and Schapire}{Freund and
  Schapire}{1995}]%
        {freund1995desicion}
\bibfield{author}{\bibinfo{person}{Yoav Freund} {and}
  \bibinfo{person}{Robert~E. Schapire}.} \bibinfo{year}{1995}\natexlab{}.
\newblock \showarticletitle{A desicion-theoretic generalization of on-line
  learning and an application to boosting}. In \bibinfo{booktitle}{{\em
  Computational learning theory}}. Springer, \bibinfo{pages}{23--37}.
\newblock


\bibitem[\protect\citeauthoryear{Jiang, Niu, Guo, Mustafa, Lin, Chen, and
  Zhou}{Jiang et~al\mbox{.}}{2013}]%
        {jiang2013novel}
\bibfield{author}{\bibinfo{person}{Xiaotian Jiang}, \bibinfo{person}{Zhendong
  Niu}, \bibinfo{person}{Jiamin Guo}, \bibinfo{person}{Ghulam Mustafa},
  \bibinfo{person}{Zihan Lin}, \bibinfo{person}{Baomi Chen}, {and}
  \bibinfo{person}{Qian Zhou}.} \bibinfo{year}{2013}\natexlab{}.
\newblock \showarticletitle{Novel Boosting Frameworks to Improve the
  Performance of Collaborative Filtering}. In \bibinfo{booktitle}{{\em
  ACML'13}}. \bibinfo{pages}{87--99}.
\newblock


\bibitem[\protect\citeauthoryear{Koren}{Koren}{2008}]%
        {DBLP:conf/kdd/Koren08}
\bibfield{author}{\bibinfo{person}{Yehuda Koren}.}
  \bibinfo{year}{2008}\natexlab{}.
\newblock \showarticletitle{Factorization meets the neighborhood: a
  multifaceted collaborative filtering model}. In \bibinfo{booktitle}{{\em
  Proceedings of the 14th {ACM} {SIGKDD} International Conference on Knowledge
  Discovery and Data Mining, Las Vegas, Nevada, USA, August 24-27, 2008}}.
  \bibinfo{pages}{426--434}.
\newblock
\showDOI{%
\url{http://dx.doi.org/10.1145/1401890.1401944}}


\bibitem[\protect\citeauthoryear{Koren}{Koren}{2010}]%
        {DBLP:journals/cacm/Koren10}
\bibfield{author}{\bibinfo{person}{Yehuda Koren}.}
  \bibinfo{year}{2010}\natexlab{}.
\newblock \showarticletitle{Collaborative filtering with temporal dynamics}.
\newblock \bibinfo{journal}{{\em Commun. {ACM}\/}} \bibinfo{volume}{53},
  \bibinfo{number}{4} (\bibinfo{year}{2010}), \bibinfo{pages}{89--97}.
\newblock
\showDOI{%
\url{http://dx.doi.org/10.1145/1721654.1721677}}


\bibitem[\protect\citeauthoryear{Liu}{Liu}{2011}]%
        {DBLP:books/daglib/0027504}
\bibfield{author}{\bibinfo{person}{Tie{-}Yan Liu}.}
  \bibinfo{year}{2011}\natexlab{}.
\newblock \bibinfo{booktitle}{{\em Learning to Rank for Information
  Retrieval}}.
\newblock \bibinfo{publisher}{Springer}.
\newblock
\showISBNx{978-3-642-14266-6}
\showDOI{%
\url{http://dx.doi.org/10.1007/978-3-642-14267-3}}


\bibitem[\protect\citeauthoryear{Liu, Zhao, Sun, and Miao}{Liu
  et~al\mbox{.}}{2015}]%
        {DBLP:conf/ijcai/LiuZSM15}
\bibfield{author}{\bibinfo{person}{Yong Liu}, \bibinfo{person}{Peilin Zhao},
  \bibinfo{person}{Aixin Sun}, {and} \bibinfo{person}{Chunyan Miao}.}
  \bibinfo{year}{2015}\natexlab{}.
\newblock \showarticletitle{A Boosting Algorithm for Item Recommendation with
  Implicit Feedback}. In \bibinfo{booktitle}{{\em Proceedings of the
  Twenty-Fourth International Joint Conference on Artificial Intelligence,
  {IJCAI} 2015, Buenos Aires, Argentina, July 25-31, 2015}}.
  \bibinfo{pages}{1792--1798}.
\newblock
\showURL{%
\url{http://ijcai.org/Abstract/15/255}}


\bibitem[\protect\citeauthoryear{McFee and Lanckriet}{McFee and
  Lanckriet}{2010}]%
        {DBLP:conf/icml/McFeeL10}
\bibfield{author}{\bibinfo{person}{Brian McFee} {and} \bibinfo{person}{Gert
  R.~G. Lanckriet}.} \bibinfo{year}{2010}\natexlab{}.
\newblock \showarticletitle{Metric Learning to Rank}. In
  \bibinfo{booktitle}{{\em Proceedings of the 27th International Conference on
  Machine Learning (ICML-10), June 21-24, 2010, Haifa, Israel}}.
  \bibinfo{pages}{775--782}.
\newblock
\showURL{%
\url{http://www.icml2010.org/papers/504.pdf}}


\bibitem[\protect\citeauthoryear{Qiang, Liang, and Yang}{Qiang
  et~al\mbox{.}}{2013}]%
        {DBLP:conf/cikm/QiangLY13}
\bibfield{author}{\bibinfo{person}{Runwei Qiang}, \bibinfo{person}{Feng Liang},
  {and} \bibinfo{person}{Jianwu Yang}.} \bibinfo{year}{2013}\natexlab{}.
\newblock \showarticletitle{Exploiting ranking factorization machines for
  microblog retrieval}. In \bibinfo{booktitle}{{\em 22nd {ACM} International
  Conference on Information and Knowledge Management, CIKM'13, San Francisco,
  CA, USA, October 27 - November 1, 2013}}. \bibinfo{pages}{1783--1788}.
\newblock
\showDOI{%
\url{http://dx.doi.org/10.1145/2505515.2505648}}


\bibitem[\protect\citeauthoryear{Rendle}{Rendle}{2010}]%
        {DBLP:conf/icdm/Rendle10}
\bibfield{author}{\bibinfo{person}{Steffen Rendle}.}
  \bibinfo{year}{2010}\natexlab{}.
\newblock \showarticletitle{Factorization Machines}. In
  \bibinfo{booktitle}{{\em {ICDM} 2010, The 10th {IEEE} International
  Conference on Data Mining, Sydney, Australia, 14-17 December 2010}}.
  \bibinfo{pages}{995--1000}.
\newblock
\showDOI{%
\url{http://dx.doi.org/10.1109/ICDM.2010.127}}


\bibitem[\protect\citeauthoryear{Rendle}{Rendle}{2012}]%
        {DBLP:journals/tist/Rendle12}
\bibfield{author}{\bibinfo{person}{Steffen Rendle}.}
  \bibinfo{year}{2012}\natexlab{}.
\newblock \showarticletitle{Factorization Machines with libFM}.
\newblock \bibinfo{journal}{{\em {ACM} {TIST}\/}} \bibinfo{volume}{3},
  \bibinfo{number}{3} (\bibinfo{year}{2012}), \bibinfo{pages}{57}.
\newblock
\showDOI{%
\url{http://dx.doi.org/10.1145/2168752.2168771}}


\bibitem[\protect\citeauthoryear{Rendle}{Rendle}{2013}]%
        {DBLP:journals/pvldb/Rendle13}
\bibfield{author}{\bibinfo{person}{Steffen Rendle}.}
  \bibinfo{year}{2013}\natexlab{}.
\newblock \showarticletitle{Scaling Factorization Machines to Relational Data}.
\newblock \bibinfo{journal}{{\em {PVLDB}\/}} \bibinfo{volume}{6},
  \bibinfo{number}{5} (\bibinfo{year}{2013}), \bibinfo{pages}{337--348}.
\newblock
\showURL{%
\url{http://www.vldb.org/pvldb/vol6/p337-rendle.pdf}}


\bibitem[\protect\citeauthoryear{Rendle, Gantner, Freudenthaler, and
  Schmidt{-}Thieme}{Rendle et~al\mbox{.}}{2011}]%
        {DBLP:conf/sigir/RendleGFS11}
\bibfield{author}{\bibinfo{person}{Steffen Rendle}, \bibinfo{person}{Zeno
  Gantner}, \bibinfo{person}{Christoph Freudenthaler}, {and}
  \bibinfo{person}{Lars Schmidt{-}Thieme}.} \bibinfo{year}{2011}\natexlab{}.
\newblock \showarticletitle{Fast context-aware recommendations with
  factorization machines}. In \bibinfo{booktitle}{{\em Proceeding of the 34th
  International {ACM} {SIGIR} Conference on Research and Development in
  Information Retrieval, {SIGIR} 2011, Beijing, China, July 25-29, 2011}}.
  \bibinfo{pages}{635--644}.
\newblock
\showDOI{%
\url{http://dx.doi.org/10.1145/2009916.2010002}}


\bibitem[\protect\citeauthoryear{Rendle and Schmidt{-}Thieme}{Rendle and
  Schmidt{-}Thieme}{2010}]%
        {DBLP:conf/wsdm/RendleS10}
\bibfield{author}{\bibinfo{person}{Steffen Rendle} {and} \bibinfo{person}{Lars
  Schmidt{-}Thieme}.} \bibinfo{year}{2010}\natexlab{}.
\newblock \showarticletitle{Pairwise interaction tensor factorization for
  personalized tag recommendation}. In \bibinfo{booktitle}{{\em Proceedings of
  the Third International Conference on Web Search and Web Data Mining, {WSDM}
  2010, New York, NY, USA, February 4-6, 2010}}. \bibinfo{pages}{81--90}.
\newblock
\showDOI{%
\url{http://dx.doi.org/10.1145/1718487.1718498}}


\bibitem[\protect\citeauthoryear{Schapire and Singer}{Schapire and
  Singer}{1999}]%
        {schapire1999improved}
\bibfield{author}{\bibinfo{person}{Robert~E. Schapire} {and}
  \bibinfo{person}{Yoram Singer}.} \bibinfo{year}{1999}\natexlab{}.
\newblock \showarticletitle{Improved boosting algorithms using confidence-rated
  predictions}.
\newblock \bibinfo{journal}{{\em Machine learning\/}} \bibinfo{volume}{37},
  \bibinfo{number}{3} (\bibinfo{year}{1999}), \bibinfo{pages}{297--336}.
\newblock


\bibitem[\protect\citeauthoryear{Srebro, Rennie, and Jaakkola}{Srebro
  et~al\mbox{.}}{2004}]%
        {DBLP:conf/nips/SrebroRJ04}
\bibfield{author}{\bibinfo{person}{Nathan Srebro}, \bibinfo{person}{Jason D.~M.
  Rennie}, {and} \bibinfo{person}{Tommi~S. Jaakkola}.}
  \bibinfo{year}{2004}\natexlab{}.
\newblock \showarticletitle{Maximum-Margin Matrix Factorization}. In
  \bibinfo{booktitle}{{\em Advances in Neural Information Processing Systems 17
  [Neural Information Processing Systems, {NIPS} 2004, December 13-18, 2004,
  Vancouver, British Columbia, Canada]}}. \bibinfo{pages}{1329--1336}.
\newblock
\showURL{%
\url{http://papers.nips.cc/paper/2655-maximum-margin-matrix-factorization}}


\bibitem[\protect\citeauthoryear{Wang, Sun, and Zhang}{Wang
  et~al\mbox{.}}{2014}]%
        {wang2014adamf}
\bibfield{author}{\bibinfo{person}{Yanghao Wang}, \bibinfo{person}{Hailong
  Sun}, {and} \bibinfo{person}{Richong Zhang}.}
  \bibinfo{year}{2014}\natexlab{}.
\newblock \showarticletitle{AdaMF: Adaptive Boosting Matrix Factorization for
  Recommender System}.
\newblock In \bibinfo{booktitle}{{\em WAIM'14}}. \bibinfo{publisher}{Springer},
  \bibinfo{pages}{43--54}.
\newblock


\bibitem[\protect\citeauthoryear{Xu and Li}{Xu and Li}{2007}]%
        {xu2007adarank}
\bibfield{author}{\bibinfo{person}{Jun Xu} {and} \bibinfo{person}{Hang Li}.}
  \bibinfo{year}{2007}\natexlab{}.
\newblock \showarticletitle{AdaRank: a boosting algorithm for information
  retrieval}. In \bibinfo{booktitle}{{\em SIGIR'07}}. ACM,
  \bibinfo{pages}{391--398}.
\newblock


\bibitem[\protect\citeauthoryear{Yuan, Guo, Jose, Chen, Yu, and Zhang}{Yuan
  et~al\mbox{.}}{2016}]%
        {DBLP:conf/cikm/YuanGJCYZ16}
\bibfield{author}{\bibinfo{person}{Fajie Yuan}, \bibinfo{person}{Guibing Guo},
  \bibinfo{person}{Joemon~M. Jose}, \bibinfo{person}{Long Chen},
  \bibinfo{person}{Haitao Yu}, {and} \bibinfo{person}{Weinan Zhang}.}
  \bibinfo{year}{2016}\natexlab{}.
\newblock \showarticletitle{LambdaFM: Learning Optimal Ranking with
  Factorization Machines Using Lambda Surrogates}. In \bibinfo{booktitle}{{\em
  Proceedings of the 25th {ACM} International on Conference on Information and
  Knowledge Management, {CIKM} 2016, Indianapolis, IN, USA, October 24-28,
  2016}}. \bibinfo{pages}{227--236}.
\newblock
\showDOI{%
\url{http://dx.doi.org/10.1145/2983323.2983758}}


\end{thebibliography}


\begin{thebibliography}{10}

\bibitem{DBLP:conf/nips/BurgesRL06}
Christopher J.~C. Burges, Robert Ragno, and Quoc~Viet Le.
\newblock Learning to rank with nonsmooth cost functions.
\newblock In {\em Advances in Neural Information Processing Systems 19,
  Proceedings of the Twentieth Annual Conference on Neural Information
  Processing Systems, Vancouver, British Columbia, Canada, December 4-7, 2006},
  pages 193--200, 2006.

\bibitem{cheng2014gradient}
Chen Cheng, Fen Xia, Tong Zhang, Irwin King, and Michael~R. Lyu.
\newblock Gradient boosting factorization machines.
\newblock In {\em RecSys'14}, pages 265--272. ACM, 2014.

\bibitem{DBLP:conf/nips/CortesM03}
Corinna Cortes and Mehryar Mohri.
\newblock {AUC} optimization vs. error rate minimization.
\newblock In {\em Advances in Neural Information Processing Systems 16 [Neural
  Information Processing Systems, {NIPS} 2003, December 8-13, 2003, Vancouver
  and Whistler, British Columbia, Canada]}, pages 313--320, 2003.

\bibitem{DBLP:conf/recsys/CremonesiKT10}
Paolo Cremonesi, Yehuda Koren, and Roberto Turrin.
\newblock Performance of recommender algorithms on top-n recommendation tasks.
\newblock In {\em Proceedings of the 2010 {ACM} Conference on Recommender
  Systems, RecSys 2010, Barcelona, Spain, September 26-30, 2010}, pages 39--46,
  2010.

\bibitem{DBLP:books/daglib/0026018}
Nello Cristianini and John Shawe{-}Taylor.
\newblock {\em An Introduction to Support Vector Machines and Other
  Kernel-based Learning Methods}.
\newblock Cambridge University Press, 2010.

\bibitem{duffy2002boosting}
Nigel Duffy and David Helmbold.
\newblock Boosting methods for regression.
\newblock {\em Machine Learning}, 47(2-3):153--200, 2002.

\bibitem{DBLP:journals/jmlr/FreundISS03}
Yoav Freund, Raj~D. Iyer, Robert~E. Schapire, and Yoram Singer.
\newblock An efficient boosting algorithm for combining preferences.
\newblock {\em Journal of Machine Learning Research}, 4:933--969, 2003.

\bibitem{freund1995desicion}
Yoav Freund and Robert~E. Schapire.
\newblock A desicion-theoretic generalization of on-line learning and an
  application to boosting.
\newblock In {\em Computational learning theory}, pages 23--37. Springer, 1995.

\bibitem{jiang2013novel}
Xiaotian Jiang, Zhendong Niu, Jiamin Guo, Ghulam Mustafa, Zihan Lin, Baomi
  Chen, and Qian Zhou.
\newblock Novel boosting frameworks to improve the performance of collaborative
  filtering.
\newblock In {\em ACML'13}, pages 87--99, 2013.

\bibitem{DBLP:conf/kdd/Koren08}
Yehuda Koren.
\newblock Factorization meets the neighborhood: a multifaceted collaborative
  filtering model.
\newblock In {\em Proceedings of the 14th {ACM} {SIGKDD} International
  Conference on Knowledge Discovery and Data Mining, Las Vegas, Nevada, USA,
  August 24-27, 2008}, pages 426--434, 2008.

\bibitem{DBLP:journals/cacm/Koren10}
Yehuda Koren.
\newblock Collaborative filtering with temporal dynamics.
\newblock {\em Commun. {ACM}}, 53(4):89--97, 2010.

\bibitem{DBLP:books/daglib/0027504}
Tie{-}Yan Liu.
\newblock {\em Learning to Rank for Information Retrieval}.
\newblock Springer, 2011.

\bibitem{DBLP:conf/ijcai/LiuZSM15}
Yong Liu, Peilin Zhao, Aixin Sun, and Chunyan Miao.
\newblock A boosting algorithm for item recommendation with implicit feedback.
\newblock In {\em Proceedings of the Twenty-Fourth International Joint
  Conference on Artificial Intelligence, {IJCAI} 2015, Buenos Aires, Argentina,
  July 25-31, 2015}, pages 1792--1798, 2015.

\bibitem{DBLP:conf/icml/McFeeL10}
Brian McFee and Gert R.~G. Lanckriet.
\newblock Metric learning to rank.
\newblock In {\em Proceedings of the 27th International Conference on Machine
  Learning (ICML-10), June 21-24, 2010, Haifa, Israel}, pages 775--782, 2010.

\bibitem{DBLP:conf/cikm/QiangLY13}
Runwei Qiang, Feng Liang, and Jianwu Yang.
\newblock Exploiting ranking factorization machines for microblog retrieval.
\newblock In {\em 22nd {ACM} International Conference on Information and
  Knowledge Management, CIKM'13, San Francisco, CA, USA, October 27 - November
  1, 2013}, pages 1783--1788, 2013.

\bibitem{DBLP:conf/icdm/Rendle10}
Steffen Rendle.
\newblock Factorization machines.
\newblock In {\em {ICDM} 2010, The 10th {IEEE} International Conference on Data
  Mining, Sydney, Australia, 14-17 December 2010}, pages 995--1000, 2010.

\bibitem{DBLP:journals/tist/Rendle12}
Steffen Rendle.
\newblock Factorization machines with libfm.
\newblock {\em {ACM} {TIST}}, 3(3):57, 2012.

\bibitem{DBLP:journals/pvldb/Rendle13}
Steffen Rendle.
\newblock Scaling factorization machines to relational data.
\newblock {\em {PVLDB}}, 6(5):337--348, 2013.

\bibitem{DBLP:conf/sigir/RendleGFS11}
Steffen Rendle, Zeno Gantner, Christoph Freudenthaler, and Lars
  Schmidt{-}Thieme.
\newblock Fast context-aware recommendations with factorization machines.
\newblock In {\em Proceeding of the 34th International {ACM} {SIGIR} Conference
  on Research and Development in Information Retrieval, {SIGIR} 2011, Beijing,
  China, July 25-29, 2011}, pages 635--644, 2011.

\bibitem{DBLP:conf/wsdm/RendleS10}
Steffen Rendle and Lars Schmidt{-}Thieme.
\newblock Pairwise interaction tensor factorization for personalized tag
  recommendation.
\newblock In {\em Proceedings of the Third International Conference on Web
  Search and Web Data Mining, {WSDM} 2010, New York, NY, USA, February 4-6,
  2010}, pages 81--90, 2010.

\bibitem{schapire1999improved}
Robert~E. Schapire and Yoram Singer.
\newblock Improved boosting algorithms using confidence-rated predictions.
\newblock {\em Machine learning}, 37(3):297--336, 1999.

\bibitem{DBLP:conf/nips/SrebroRJ04}
Nathan Srebro, Jason D.~M. Rennie, and Tommi~S. Jaakkola.
\newblock Maximum-margin matrix factorization.
\newblock In {\em Advances in Neural Information Processing Systems 17 [Neural
  Information Processing Systems, {NIPS} 2004, December 13-18, 2004, Vancouver,
  British Columbia, Canada]}, pages 1329--1336, 2004.

\bibitem{wang2014adamf}
Yanghao Wang, Hailong Sun, and Richong Zhang.
\newblock Adamf: Adaptive boosting matrix factorization for recommender system.
\newblock In {\em WAIM'14}, pages 43--54. Springer, 2014.

\bibitem{xu2007adarank}
Jun Xu and Hang Li.
\newblock Adarank: a boosting algorithm for information retrieval.
\newblock In {\em SIGIR'07}, pages 391--398. ACM, 2007.

\bibitem{DBLP:conf/cikm/YuanGJCYZ16}
Fajie Yuan, Guibing Guo, Joemon~M. Jose, Long Chen, Haitao Yu, and Weinan
  Zhang.
\newblock Lambdafm: Learning optimal ranking with factorization machines using
  lambda surrogates.
\newblock In {\em Proceedings of the 25th {ACM} International on Conference on
  Information and Knowledge Management, {CIKM} 2016, Indianapolis, IN, USA,
  October 24-28, 2016}, pages 227--236, 2016.

\end{thebibliography}

\end{document}